\documentclass[runningheads]{llncs}

\usepackage{eccv}

\usepackage{eccvabbrv}

\usepackage{graphicx}
\usepackage{booktabs}
\usepackage{multirow}
\usepackage{enumitem}

\usepackage[accsupp]{axessibility}  % Improves PDF readability for those with disabilities.

\usepackage[pagebackref,breaklinks,colorlinks,citecolor=eccvblue]{hyperref}
\usepackage{orcidlink}

\begin{document}

\title{A$^{2}$-MAE: A Spatial-temporal-spectral Unified Remote Sensing Pre-training Method Based on Anchor-aware Masked Autoencoder} 

\titlerunning{A$^{2}$-MAE}

\author{Lixian Zhang\inst{1,2,\thanks{Those authors contribute equally to this work.}} \and Yi Zhao\inst{2,\footnotemark[1]} \and Runmin Dong\inst{2, \footnotemark[1], \thanks{Corresponding authors.}} \and Jinxiao Zhang\inst{2} \and Shuai Yuan\inst{3} \and Shilei Cao\inst{4} \and Mengxuan Chen\inst{2} \and Juepeng Zheng\inst{4,\footnotemark[2]} \and Weijia Li\inst{4} \and Wayne Zhang\inst{5} \and Wei Liu\inst{5} \and Litong Feng\inst{5} \and Haohuan Fu\inst{2,\footnotemark[2]}}

\authorrunning{Zhang, L. et al.}

\institute{$^{1}$National Supercomputing Center in Shenzhen \quad
$^{2}$Tsinghua University \quad \\
$^{3}$The University of Hong Kong \quad
$^{4}$Sun Yat-Sen University \quad
$^{5}$SenseTime Group
{\tt\small zhanglx18@tsinghua.org.cn}~
{\tt\small \{drm,haohuan\}@mail.tsinghua.edu.cn}
}

\maketitle

\begin{abstract}
Vast amounts of remote sensing (RS) data provide Earth observations across multiple dimensions, encompassing critical spatial, temporal, and spectral information which is essential for addressing global-scale challenges such as land use monitoring, disaster prevention, and environmental change mitigation. Despite various pre-training methods tailored to the characteristics of RS data, a key limitation persists: \textit{the inability to effectively integrate spatial, temporal, and spectral information within a single unified model}. To unlock the potential of RS data, we construct a Spatial-Temporal-Spectral Structured Dataset (STSSD) characterized by the incorporation of multiple RS sources, diverse coverage, unified locations within image sets, and heterogeneity within images. Building upon this structured dataset, we propose an Anchor-Aware Masked AutoEncoder method (A$^{2}$-MAE), leveraging intrinsic complementary information from the different kinds of images (featuring different resolutions, spectral compositions, and acquisition times) and geo-information to reconstruct the masked patches during the pre-training phase. A$^{2}$-MAE integrates an anchor-aware masking strategy and a geographic encoding module to comprehensively exploit the properties of RS images. Specifically, the proposed anchor-aware masking strategy dynamically adapts the masking process based on the meta-information of a pre-selected anchor image, thereby facilitating the training on images captured by diverse types of RS sources within one model. Furthermore, we propose a geographic encoding method to leverage accurate spatial patterns, enhancing the model generalization capabilities for downstream applications that are generally location-related. Extensive experiments demonstrate our method achieves comprehensive improvements across various downstream tasks compared with existing RS pre-training methods, including image classification, semantic segmentation, and change detection tasks. The dataset and pre-training model will be released.
\end{abstract}

\begin{figure}[t]
    \centering
    \includegraphics[width=\linewidth]{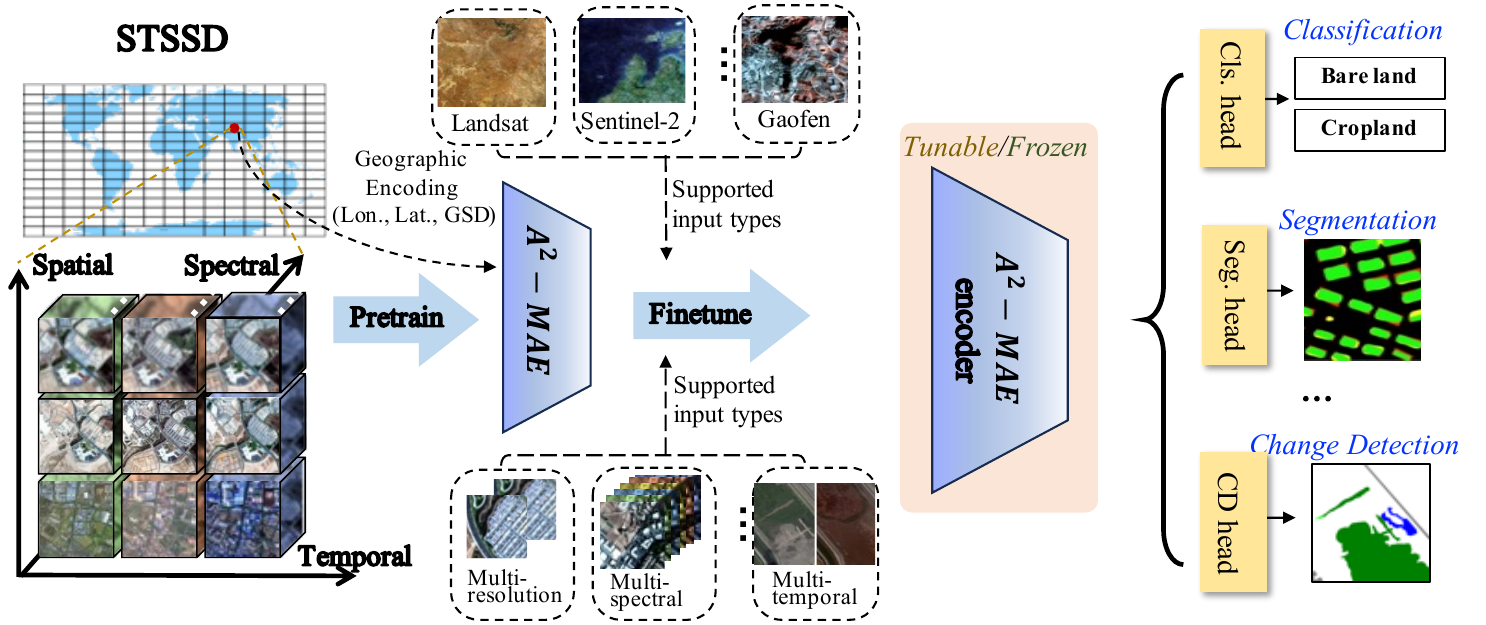}
    \caption{An illustrative overlook of the proposed global-scale dataset STSSD and pre-training method A$^{2}$-MAE. STSSD is a comprehensive remote sensing dataset structured and characterized by the inclusion of diverse spatial, temporal, and spectral coverage. A$^{2}$-MAE facilitates the efficient utilization of the intrinsic complementarity information in STSSD within one unified spatial-temporal-spectral model. Lon., Lat., and GSD indicate longitude, latitude, and ground spatial distance information, respectively.}
    \label{fig:front}
    \vspace{-0.2cm}
\end{figure}

\section{Introduction}
\label{sec:intro}

Earth observations through remote sensing (RS) constitute a fundamental tool for monitoring the evolution of global-scale phenomena, including the urbanization process \cite{RN19, RN16}, land-use change \cite{RN76, RN75}, and biodiversity loss \cite{turner2014sensing, wang2019remote}. Over the past half-century, there has been a substantial increase in the volume of RS data, resulting in spatial, temporal, and spectral diversities within extensive RS image archives. The spatial-temporal-spectral diversities inherent in RS images offer critical and complementary information for comprehensive analysis and recognition of objects and scenes. Consequently, RS plays an pivotal role in various operational and complex research domains within the field of geoscience.

%In light of challenges faced by massive RS downstream applications due to a paucity of costly annotations \cite{wang2022self}, self-supervised learning (SSL) \cite{he2020moco,he2022masked} has emerged as a promising technique for acquiring robust feature representations from an extensive repository of satellite images. The learned feature representation can be fine-tuned with limited labels for specific downstream applications. Despite efforts in developing large pre-trained models using masked autoencoders (MAE) or contrastive learning (CL), existing RS SSL methods are tailored for specific scenarios, such as temporal SatMAE \cite{cong2022satmae}, multi-spectral SatMAE \cite{cong2022satmae}, scale-MAE\cite{reed2023scalemae}, and spatiotemporal foundation model\cite{yao2023ringmo}.
%a limited range of spatial-temporal-spectral structures and the representativeness of RS data. 
%Consequently, there are limitations in the mining and interpretation abilities of the learned representations when applied to extensive and varied unstructured spatial-temporal-spectral RS data. In this work, we pose the question: \textit{how can we leverage diverse RS images to establish a spatial-temporal-spectral-unified RS foundation model on a global scale?}

In response to the challenges encountered by extensive RS downstream applications arising from the scarcity of expensive annotations \cite{wang2022self}, self-supervised learning (SSL) \cite{he2020moco,he2022masked} has emerged as a promising technique for deriving robust feature representations from a vast repository of satellite images. The acquired feature representations can subsequently be fine-tuned with limited labeled data for specific downstream applications. Despite considerable efforts in constructing large pre-trained models through methods like masked autoencoders (MAE) or contrastive learning (CL), most of the existing RS SSL methodologies have been custom-tailored for specific scenarios, such as temporal SatMAE \cite{cong2022satmae}, multi-spectral SatMAE \cite{cong2022satmae}, multi-resolution ScaleMAE \cite{reed2023scalemae}, and spatiotemporal foundation models \cite{yao2023ringmo}. These methods enhance performance in specific downstream tasks but fall short of achieving comprehensive improvements across various downstream tasks. Besides, the SSL methods underutilize geographical information, a powerful prior for leveraging spatial patterns. In the work, we address the pivotal question: \textit{How can we present a single spatial-temporal-spectral unified RS pre-training method to effectively leverage a diverse collection of RS images?}

%exhibit limitations in handling the diverse spatial-temporal-spectral structures inherent in extensive and varied unstructured RS data, resulting in constraints on the mining and interpretation capabilities of the learned representations. 

The key to this question lies in two aspects. The first aspect involves the construction of a location-unified and extensive RS dataset encompassing images with varying temporal, spatial resolutions, and spectral compositions. Presently available RS datasets are typically derived from one or two satellite sources, offering limited spatial-temporal-spectral coverage. For instance, the Million-AID dataset exclusively covers optical RS images with RGB bands \cite{Long2021DiRS,Long2022ASP}. SEN12MS \cite{schmitt2019sen12ms} and SSL4EO-S12 \cite{wang2022ssl4eo} exhibit constrained temporal coverage. The SeCo \cite{manas2021seco} and CACo \cite{mall2023caco} datasets are pre-trained exclusively on Sentinel-2 images. However, real-world RS data exhibits significant variations in spatial resolution, temporal coverage, and spectral composition. The SSL models trained on homogeneous RS data struggle to provide effective representation for fine-tuning of downstream tasks involving different RS sources. %Therefore, mining the intrinsic relevance within a dataset featuring images with varying temporal, spatial resolution, and spectral composition at identical locations contributes to enhance the model representation.

The second aspect entails mining the intrinsic relevance from the images with different spatial, temporal, and spectral characteristics through SSL techniques. A straightforward approach is to design separate backbones for different types of sources and align the representations of different types \cite{guo2023skysense}. However, this method leads to a linear escalation in model parameters and computational overhead with the expansion of source type count. As there are a large number of types of RS sources, such as Landsat-8 with 7 bands, Sentinel-2 with 13 bands, Gaofen-2 with 4 bands, and WorldView-2 with 8 bands, it is difficult to simultaneously model the relationship across different types of sources.

%Existing RS SSL approaches are constrained by unstructured RS datasets. Bridging the performance gap among existing RS SSL approaches requires leveraging diverse spatial-temporal-spectral structures of RS data in the evolution of SSL techniques.

To address these challenges, we introduce STSSD (Figure \ref{fig:front}), a global-scale RS dataset containing half a million sampling locations with 2.5 million spatial-temporal-spectral structured image sets collected from multiple multi-spectral sources. Each image set is meticulously crafted to exhibit different spatial resolutions, temporal and spectral compositions for the same location. Our data processing method preserves heterogeneity within images and diversity across images for SSL. To harness the rich and varied representation features within STSSD effectively, we propose an Anchor-Aware Masked AutoEncoder method (A$^{2}$-MAE), including an anchor-aware masking strategy and geographic encoding module (Figure \ref{fig:front}). The proposed anchor-aware masking strategy enables training on images captured by diverse sources within one unified spatial-temporal-spectral model. Besides, the proposed geographic encoding method allows the model to leverage accurate spatial patterns, unleashing the potential of geo-location priors for downstream tasks. Experiments verify that our method achieves comprehensive improvements across various downstream tasks compared with state-of-the-art RS SSL methods. Taking DynamicEarthNet as an example, the performance can be further enhanced by over 8.4\% on mIoU through the introduction of geographic information during the fine-tuning process (refer to Sec. \ref{sec:ablation}).

%Leveraging the learned spatial-temporal-spectral representative features, we validate the effectiveness of A$^{2}$-MAE by adaptively fine-tuning it for diverse downstream tasks.

In summary, our contributions are as follows:

\begin{itemize}[label=\textbullet, leftmargin=1em]
\item We build the STSSD, a globally spatial-temporal-spectral structured RS dataset featuring high diversity, unification, and heterogeneity. STSSD is meticulously curated to encompass diverse land-use types spatially, capture landscape changes temporally, and incorporate various band compositions spectrally.

%a comprehensive spatial-temporal-spectral unified remote sensing dataset comprising over 4.2 million RS images. STSSD is meticulously curated to encompass diverse land-use types spatially, capture landscape changes temporally, and incorporate various band compositions spectrally.

\item We propose a pre-training method, A$^{2}$-MAE, designed to accommodate various types of RS sources within a unified backbone architecture. A$^{2}$-MAE leverages spatial-temporal-spectral relationships and geographical information to improve model representation and generalization capabilities.

%\item We present A$^{2}$-MAE, an innovative Anchor-Aware Masked AutoEncoder featuring a novel anchor-aware masking strategy and a geographic encoding module. This architectural efficiently harnesses the spatial-temporal-spectral-unified representation along with a globally-scaled geographic embedding, thereby enhancing the effectiveness and generalization of the pre-training model.
\item Experiments verify the effectiveness and advantages of A$^{2}$-MAE compared to existing RS pre-training models with similar complexities across image classification, semantic segmentation, and change detection tasks.
\end{itemize}

\section{Related Work}

\subsection{Large-scale datasets for remote sensing imagery pre-training} 
 
Inspired by the achievements of computer vision (CV) datasets \cite{deng2009imagenet,lin2014microsoft,everingham2015pascal,abu2016youtube}, researchers have introduced several large-scale RS datasets \cite{christie2018fmow,sumbul2019bigearthnet,sumbul2021bigearthnet,toker2022dynamicearthnet,wang2022ssl4eo,bastani2023satlaspretrain}. 
 
These datasets exhibit a gradual expansion in the volume of data, starting the fMoW \cite{christie2018fmow} encompassing 1 million images, progressing to BigEarthNet-MM \cite{sumbul2021bigearthnet} with 1.2 million images, and further expanding to SSL4EO-S12 \cite{wang2022ssl4eo} comprising 3 million images.
Additionally, there has been a progression in the diversity of spectral sources in datasets, transitioning from datasets like BigEarthNet \cite{sumbul2019bigearthnet} solely from Sentinel-1, to BigEarthNet-MM \cite{sumbul2021bigearthnet} combining Sentinel-1/2 pairs, to SatlasPretrain \cite{bastani2023satlaspretrain}, which incorporates data from Sentinel-1/2 and NAIP, and then to DynamicEarthNet \cite{toker2022dynamicearthnet} containing diverse spatial-temporal-spectral images with constrained sampling locations.  
% However, these datasets are tailored for specific applications (\eg DynamicEarthNet for tracking change and BigEarthNet-MM for RS image retrieval and classification). Furthermore, most of them are collected from one or two satellite sources, resulting in a lack of diversity in spatial-temporal-spectral coverage. 
Therefore, there is an urgent need to construct a large-scale spatial-temporal-spectral structured RS dataset encompassing more multi-spectral sources and diverse coverage.
%As an illustration, SSL4EO-S12 \cite{wang2022ssl4eo} and SatlasPretrain \cite{bastani2023satlaspretrain} exclusively provide global observations for the years 2021 and 2022, respectively, whereas observations in \cite{bastani2023satlaspretrain} specific to the US cover the period from 2011 to 2020.
In this work, we introduce STSSD for spatial-temporal-spectral unified learning, surpassing the DynamicEarthNet dataset by 10 times and incorporating data from 4 multi-spectral sources.

\subsection{Self-supervised learning for satellite imagery}
SSL primarily focuses on generating supervisory signals from unlabeled data, through the design of various pretext tasks such as 
% rotation prediction \cite{gidaris2018unsupervised},  colorization \cite{zhang2016colorful},   
masked patches reconstruction \cite{he2022masked,tong2022videomae,zhang2022graph,bachmann2022multimae,woo2023convnext} and contrasting semantically similar inputs \cite{chen2020simple,he2020moco,grill2020bootstrap,chen2021exploring,caron2021emerging}. 
Furthermore, SSL enables the acquisition of semantic information without human annotation.
Therefore, SSL plays a vital role in the RS domain \cite{wang2022self}, where annotation demands specialized expertise and incurs high costs. 
Existing RS pre-training methods leverage different properties of RS images or specific RS tasks \cite{wang2022self}.
For instance, Ayush et al.\cite{ayush2021geography} leverage spatially aligned but temporally separated images as positive pairs to learn feature representation for 10m multi-spectral images.
Similarly, Mall et al.\cite{mall2023caco} propose a new SSL loss for CL to distinguish between short-term and long-term changes in multi-spectral images.
% \citet{gao2022general} demonstrated the effectiveness of MAE pre-training for the classification of RS images.  
Cong et al.\cite{cong2022satmae} introduce SatMAE to leverage temporal or multi-spectral information in data through positional encoding.
Reed et al.\cite{reed2023scalemae} present Scale-MAE to reconstruct both low and high-frequency images to learn robust multi-scale representations for RS imagery. 
Nevertheless, these studies are customized for specific types of RS images and cannot simultaneously utilize RS images from different kinds of multi-spectral sources in one unified model.
%spatial-temporal-spectral representations and may not be optimally equipped to handle diverse spatial-temporal-spectral information, hindering the achievement of comprehensive representation learning.
To fill this gap, we propose an anchor-aware masking strategy to leverage intrinsic complementarity information from an image set, which can be easily extended to various multi-spectral sources.

\subsection{Geography-aware learning}
RS images offer essential metadata records containing geographic information, such as geographic location and ground sample distance (GSD) \cite{WU202187, li2021geographical}. This prior information enables the capture of geographic patterns and fosters a robust linkage between fine-tuning data and models pre-trained globally \cite{zhao2023mine,geoiccv}. Consequently, it is anticipated to bolster the representational capacity of the pre-trained model \cite{satclip, bai2023geographic}. While a few studies have leveraged recorded geographic data \cite{ayush2021geography,grocvprw}, they are constrained in efficiently utilizing such information on a large scale \cite{2023geographic}. one-hot geo-encoding \cite{christie2018fmow} offers limited encoding outcomes, while GSD scaling encoding \cite{reed2023scalemae} cannot be jointly integrated with geo-location data. Another alternative (\ie, geo-context prototype learning \cite{guo2023skysense}) demands additional computational resources while yielding encoding outcomes unsuitable for varying spatial resolutions. To bridge this gap, we introduce a Geographic Encoding Module in A$^{2}$-MAE, providing more accurate geographical priors without additional computation overhead (\ie, latitude, longitude, and GSD), thereby improving the generalization of applications on a global scale. 
%enhancing representation learning performance across worldwide RS datasets.

\section{Data}
\subsection{Overview}
\begin{figure*}[htp]
    \centering
    \includegraphics[width=\linewidth]{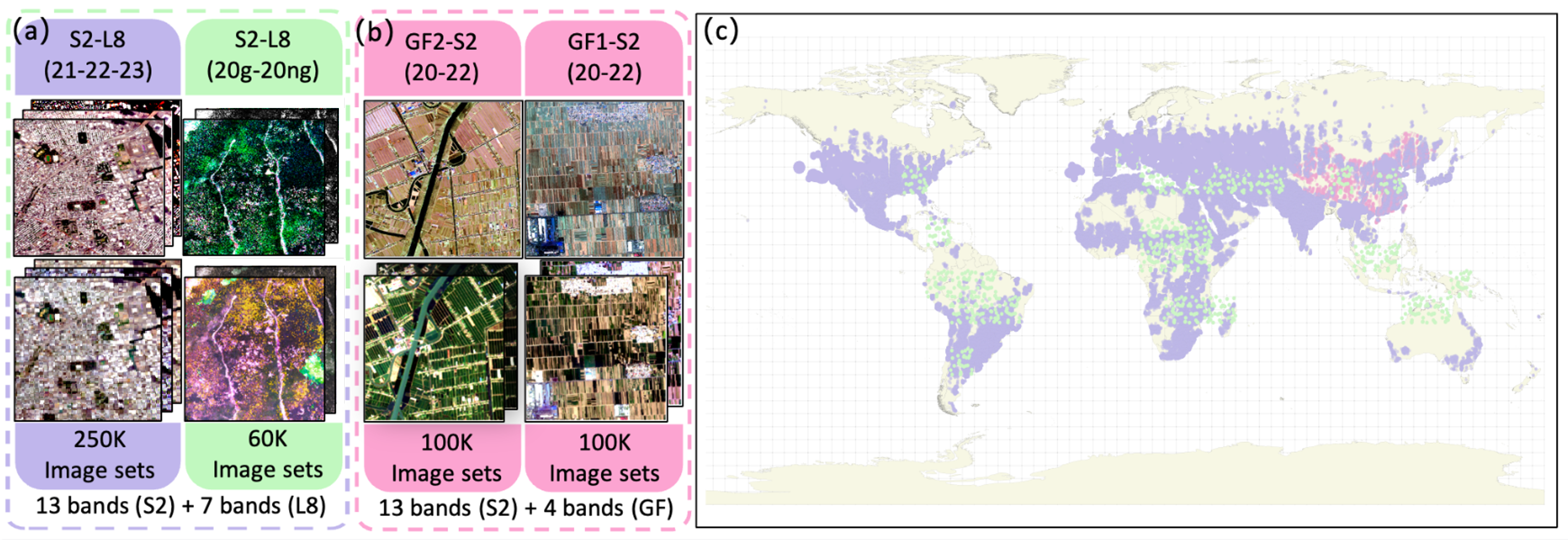}
    \caption{The compositions of image sets and the global sampling location distribution. (a) shows the S2-L8 image sets. "ng" denotes the non-growth period, and "g" denotes the growth period within one year. (b) shows the GF-S2 image sets. (c) is the sampling location distribution (i.e., purple circles for S2-L8 in urban areas, green circles for S2-L8 in nature reserves, and pink circles for GF-S2).}
    \label{fig:data}
    \vspace{-0.2cm}
\end{figure*}

We introduce STSSD, a large-scale RS dataset designed for spatial-temporal-spectral unified SSL. This dataset is meticulously curated through data pruning from an initial pool of 4.2 million original images collected at 1,045K sampling locations. The resulting STSSD comprises 510K image sets, each containing up to six images collected from two sources with different resolutions and spectral compositions.  As shown in Figure \ref{fig:data},  STSSD consists of 4 kinds of image sets, featuring diverse sources, spatial resolutions, coverage, and acquisition times. It is characterized by the following four key attributes:

%Multi-Sensor(数据源的不同波段，不同分辨率)/Unification（image set里面包含了不同分辨率和光谱，结构亮点）/coverage diversity（支撑编码，地物类型）/异质性（裁剪之后）
\textbf{Diversity.} STSSD exhibits comprehensive source diversity, containing 4 different satellite sources with 3 different band compositions. It also boasts spectral diversity, manifested in different band compositions with resolutions ranging from 0.8m/pixel to 30m/pixel, derived from various and distinct data sources.

\textbf{Coverage.} STSSD owns dynamic and diverse geographical features, involving more than 12,000 urban centers and 10,000 nature reserves around the world. This global coverage enhances the model's abilities in downstream tasks with diverse coverage and geographical characteristics.

\textbf{Unification.} STSSD integrates spatial, temporal, and spectral contexts to generate image sets sourced from different origins and acquisition times. These image sets are spatial-temporal-spectral unified, providing more dimensions of information compared to single-temporal or single-source data, thereby significantly contributing to the unification and robustness of the RS foundation model.

\textbf{Heterogeneity.} Employing a clustering-based data pruning strategy to eliminate redundant types (\eg, desert) and low-quality images, STSSD balances the diversity across images and heterogeneity within an image. The heterogeneity can increase the difficulty of image reconstruction for SSL, thereby empowering the model with enhanced feature representation capabilities.

%\textbf{Scale.} STSSD comprises over 1,045K sampling locations globally, featuring in excess of 4.2 million images.
%\textbf{Diversity.} STSSD exhibits comprehensive spatial diversity, encompassing sampling locations in major urban centers and natural reserves worldwide. It also boasts temporal diversity through periodic revisits to sampling locations, as well as spectral diversity, manifested in different band compositions with resolutions ranging from 0.8m/pixel to 30m/pixel, derived from various and distinct data sources.
%\textbf{Unification.} STSSD integrates spatial, temporal, and spectral contexts to generate image sets sourced from different origins and acquisition times. These image sets encapsulate rich temporal-spectral information, a notable enhancement over single-temporal or single-source data, thereby significantly contributing to the unification and robustness of the RS foundation model.
%\textbf{Heterogeneity.} Employing a judicious data pruning strategy to eliminate redundant types (\eg, desert) and low-quality images, STSSD balances the diversity and heterogeneity for a more curated dataset. This deliberate heterogeneity not only facilitates but also challenges the model to discern diverse patterns, thereby enhancing its capacity to navigate complex variations encountered in RS applications.%In the remainder of this section, we provide details on image pair structures and statistics of STSSD.

\subsection{STSSD Construction}

In the pursuit of constructing a unified RS dataset characterized by diverse sources, we strategically opt for Gaofen-1 (4 bands with 1 m/pixel), Gaofen-2 (4 bands with 0.8 m/pixel), Sentinel-2 (13 bands with 10 m/pixel), and Landsat-8 (7 bands with 30 m/pixel) as our sources to maximize coverage while considering sources' accessibility, each contributing diverse resolutions, with variations up to 37.5$\times$, and multiple band compositions. %Gaofen-1 contains 4 bands with a 1 meter resolution, and Gaofen-2 contains 4 bands with a 0.8 meter resolution. Sentinel-2 has 13 bands with a 10 meter resolution, and  

Building upon the texture-rich images of urban areas proved by previous work \cite{mall2023caco}, we further expand to include nature reserves \cite{turner2014sensing,wang2019remote}, so as to enhance the model's understanding and capabilities of a more diversified and dynamic planet. Consequently, we meticulously select over the original 1,045K sampling locations, spanning nature reserves (depicted in green) and main cities (depicted in purple), to collect Sentinel-2 and Landsat-8 image sets (S2-L8), as illustrated in Figure \ref{fig:data} (c). To capture the dynamic nature of geographical features, a time series of images are provided for each sampling location, ranging from the year 2020 to 2023, with periodic seasonal revisits. Furthermore, we utilize the locations of the available Gaofen images to gather the corresponding Sentinel-2 images, subsequently forming Sentinel-2 and Gaofen image sets (GF-S2) to enhance the representation ability for higher-resolution data (depicted in pink in Figure \ref{fig:data}).

The structuring of these image sets is designed to ensure optimal resolution and band gaps for effective model learning. Specifically, there are 2 kinds of image sets: S2-L8 image sets and GF-S2 image sets. For S2-L8 (Figure \ref{fig:data} (a)), the image sets collected from main cities comprise 6 images, involving 3 Sentinel-2 and 3 Landsat-8 images annually from 2021 to 2023, to capture the temporal changes in land cover. As for nature reserves, we conduct the image sets comprising 4 images, including 2 Sentinel-2 and 2 Landsat-8 images during both the growth and non-growth periods in 2020, to showcase the phenological characteristics. For GF-S2 (Figure \ref{fig:data} (b)), each image set integrates 3 images, including a Gaofen-1 or Gaofen-2 image and 2 Sentinel-2 images captured at different time points. Note that each image set comprises two distinct data sources with different sources and temporal snapshots. The integration of diverse image sources, characterized by varying spatial resolutions, within a single image set enables multi-scale observation of the same geographical areas. This simultaneous consideration of fine-grained details and broader contextual views facilitates a more comprehensive feature representation, consequently enhancing performance across diverse downstream tasks such as building extraction and land cover mapping. Moreover, the incorporation of multi-temporal information ensures accessibility to temporal dynamics, thereby fortifying the robustness of temporal variations for downstream tasks such as change detection.

Since the original STSSD contains observations from areas with high homogeneities, such as deserts which do not contribute significantly to the diversity and complexity of the dataset due to their uniform nature, we employ a data pruning strategy to remove redundant contents and filter out low-quality images, resulting in a more refined and curated collection of data. This process ensures that the images in STSSD are high-quality and heterogeneous. After pruning, the final STSSD owns over 510K sampling locations with 2.5 million curated images. Refer to supplementary material for more details about STSSD, such as the data retrieving and pre-processing.
%These image sets integrate various spatial, temporal and spectral contexts, aiming to achieve unification o the RS foundation model.  
%To enhance the quality of the dataset, targeted pre-processing steps are undertaken for each data source. Initially, all data undergoes processing for atmosphere and radiation correction. Subsequently, the Gaofen series images are pan-sharpened to achieve higher resolution. A data pruning strategy is employed to balance the diversity and heterogeneity for a more curated dataset. This strategy involves discarding samples of redundant types (\eg, desert) and those with over 10\% cloud cover or significant no-data areas, leading to varying image counts for different geographical features.

\section{Methodology}
\label{sec:blind}
\subsection{Overall Architecture}

\begin{figure*}
    \centering
    \includegraphics[width=\linewidth]{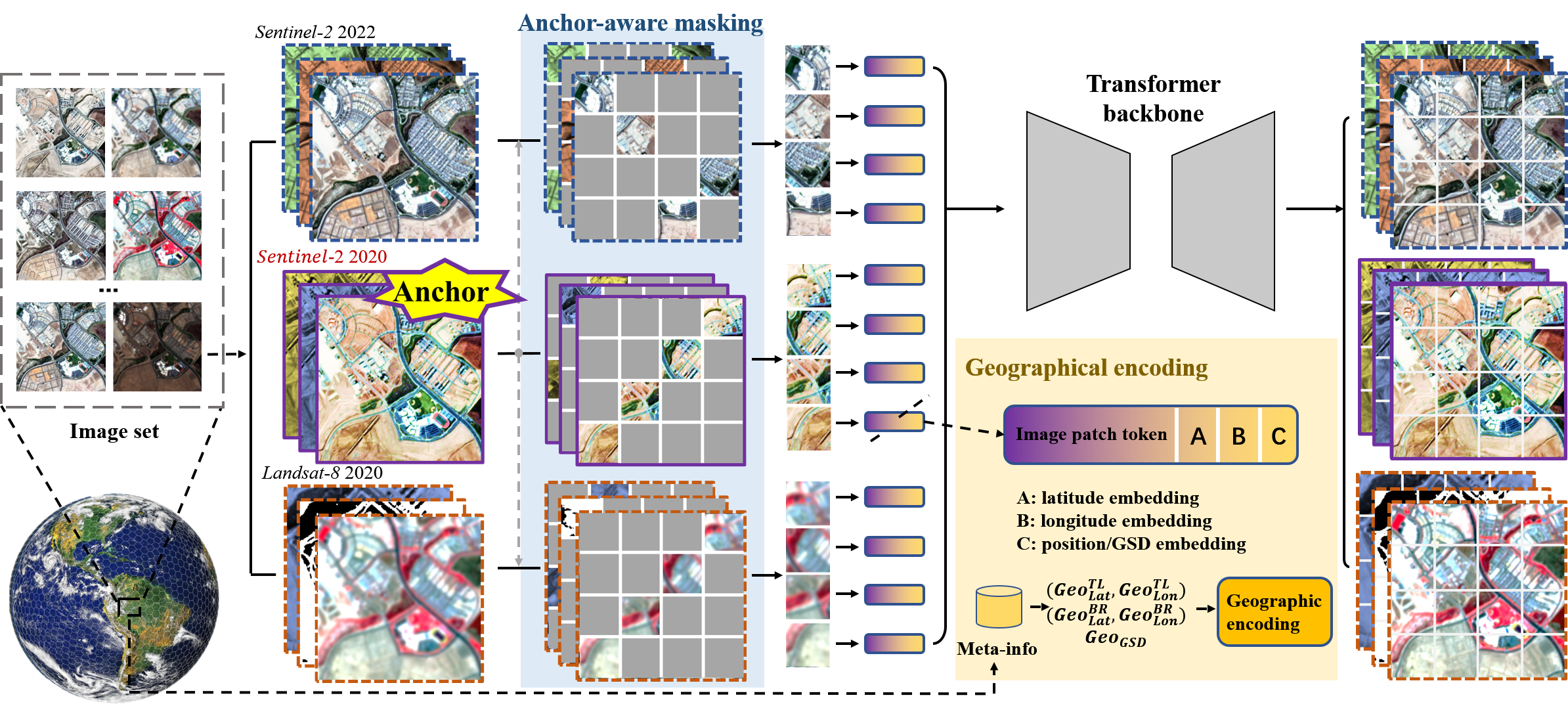}
    \caption{The overall framework of A$^{2}$-MAE. A$^{2}$-MAE incorporates an anchor-aware masking strategy and a geographic encoding module, allowing for the efficient utilization of spatial-, temporal-, and spectral-variant information in large-scale RS imagery.}
    \label{fig:overall}
    \vspace{-0.3cm}
\end{figure*}

As illustrated in Figure \ref{fig:overall}, A$^{2}$-MAE is a self-supervised pre-training method based on the MAE \cite{he2022masked}, which makes two key contributions to the MAE framework to unlock representative potentials in STSSD. First, A$^{2}$-MAE presents an anchor-aware masking (AAM) strategy to utilize the image sets collected from different sources. The AAM dynamically adjusts the masking strategy according to the meta-information of a pre-selected anchor image for each training iteration. This adaptation learning allows the model to leverage the intrinsic complementarity of spatial-temporal-spectral information to reconstruct the masked patches, thereby improving the model representation ability. In addition, A$^{2}$-MAE introduces a geographic encoding module (GEM) to obtain the geo-embedding of the given image set to provide accurate geographical priors for A$^{2}$-MAE, improving the model generalization ability.

\subsection{Setup}
Since the image sets in the STSSD are gathered by geo-locations, we denote $ P_{i} = \{I_{i}^{1, 1}, ..., I_{i}^{t, s}\} $ represents the image set at the $i^{th}$ location. $ I_{i}^{t, s} \in \mathbb {R}^{H \times W \times C} $ represents an RS image with $H$ height, $W$ width, and $C$ channels, which is captured bu source $s$ at time $t$. Note that images from different sources $s$ have different representative features, including different spectral compositions and GSDs. Three images of $P_{i}$ are randomly selected as input $I_{in}$ for A$^{2}$-MAE, ensuring a minimum of 2 different $s$ and 2 different $t$ to capture sufficient diversity in spatio-temporal-spectral relationships while balance computational cost. The A$^{2}$-MAE then patchifies the selected $I_{in}$ into three sets of sequence $Seq$ of independent patches. After randomly removing a fraction of the obtained patches, the A$^{2}$-MAE reconstructs the removed patches by leveraging the complementarity information within the remaining patches from $Seq$. Unlike the traditional MAE architecture, the A$^{2}$-MAE includes an AAM to encourage the A$^{2}$-MAE to implicitly leverage the intrinsic complementarity information within $I_{in}$ and a GEM to introduce the geographic-related information.

\subsection{Anchor-Aware Masking Strategy}

%加一段，在时间空间光谱上都需要做随机masking来支撑。但其中带来的问题是高分辨率数据会为低分辨率数据带来数据泄露，导致会出现学习捷径，所以我们设计一个策略来一致性的保证做masking来防止数据泄露带来的问题，ref to实验结果，再起一行，说结果，类似的，这个在高分和sentinel的group中也是使用这个准则。

Existing RS SSL methods are often tailored for specific scenarios, limiting the capabilities to leverage symbiotic features among images and increasing the costs for transferring to other scenarios. In contrast, our method works towards a unified pre-training method that potentially benefits various representative features between images with different spatial resolutions, temporal, and spectral compositions in $ P_{i} $. However, this symbiotic and diverse complementarity information within the image sets also poses challenges for obtaining robust and generalized RS representative features due to the complexity of spatial-temporal-spectral relationships. 

To jointly utilize the spatial-temporal-spectral information within the image sets, a straightforward method is to apply the random masking strategy to different spectral combinations of the input image set. However, when input image sets have diverse combinations of spatial resolutions and temporal compositions, the random masking strategy may lead to feature leakage from the remaining high-resolution patches during the reconstruction of the removed low-resolution patches at the same position, resulting in shortcut learning during model pre-training. To this end, we propose the AAM to dynamically adjust the masking strategy of images for each input $I_{in}$, enabling training with images from diverse sources while preventing feature leakage. Specifically, we adopt a consistent masking strategy for images from different sources $s$ at the same retrieval time $t$, a mutually-exclusive masking strategy for images from the same $s$ at different $t$, and a random masking strategy for the other circumstances. For a quantitative ablation study on AAM, please refer to Section \ref{sec:ablation}

In the example depicted in Figure \ref{fig:overall}, three images is randomly sampled from an image set $ P_{i} $, specifically, $I_{in}=\{I_{i}^{2020, Sen2}$, $I_{i}^{2020, Lan8}$, $I_{i}^{2022, Sen2}\}$. Taking the middle image  $I_{i}^{2020, Sen2}$ as the referenced anchor, A$^{2}$-MAE explicitly \textbf{\emph{aware}} the meta-information (\ie, source $s$ and time $t$) to opt specific masking strategy for removing patches from the other two images (\ie, $I_{i}^{2020, Lan8}$ and $I_{i}^{2022, Sen2}$). We first randomly select three bands of the anchor image $I_{i}^{2020, Sen2}$ to encompass a substantial diversity of band compositions while balancing the computational costs. If an image in $I_{in}$ differs in $s$ but the same in $t$ as the anchor image (\ie, the bottom image $I_{i}^{2020, Lan8}$), a consistent masking strategy is employed, obtaining a patch sequence $Seq_{i}^{2020, Lan8}$ where the patches are removed from the same position as those in $Seq_{i}^{2020, Sen2}$. This ensures that remaining patches maintain their coarsest version for the same source $s$, preventing the A$^{2}$-MAE from feature leakage during pre-training. To address temporal disparities, if an image in $I_{in}$ has a different time $t$ from the same $s$ (\ie, the upper image $I_{i}^{2022, Sen2}$), a mutually-exclusive masking strategy is adopted to ensure the removed patch retains positional uniqueness with $Seq_{i}^{2020, Sen2}$, enhancing A$^{2}$-MAE's capacity to leverage multi-temporal symbiotic features. Additionally, the incorporation of $I_{in}$ offers sufficient diversity in spatio-temporal-spectral relationships, encouraging A$^{2}$-MAE to effectively leverage multi-scale symbiotic features for patch reconstruction.

\subsection{Geographic Encoding}

\begin{figure}[t]
    \centering
    \includegraphics[width=10cm]{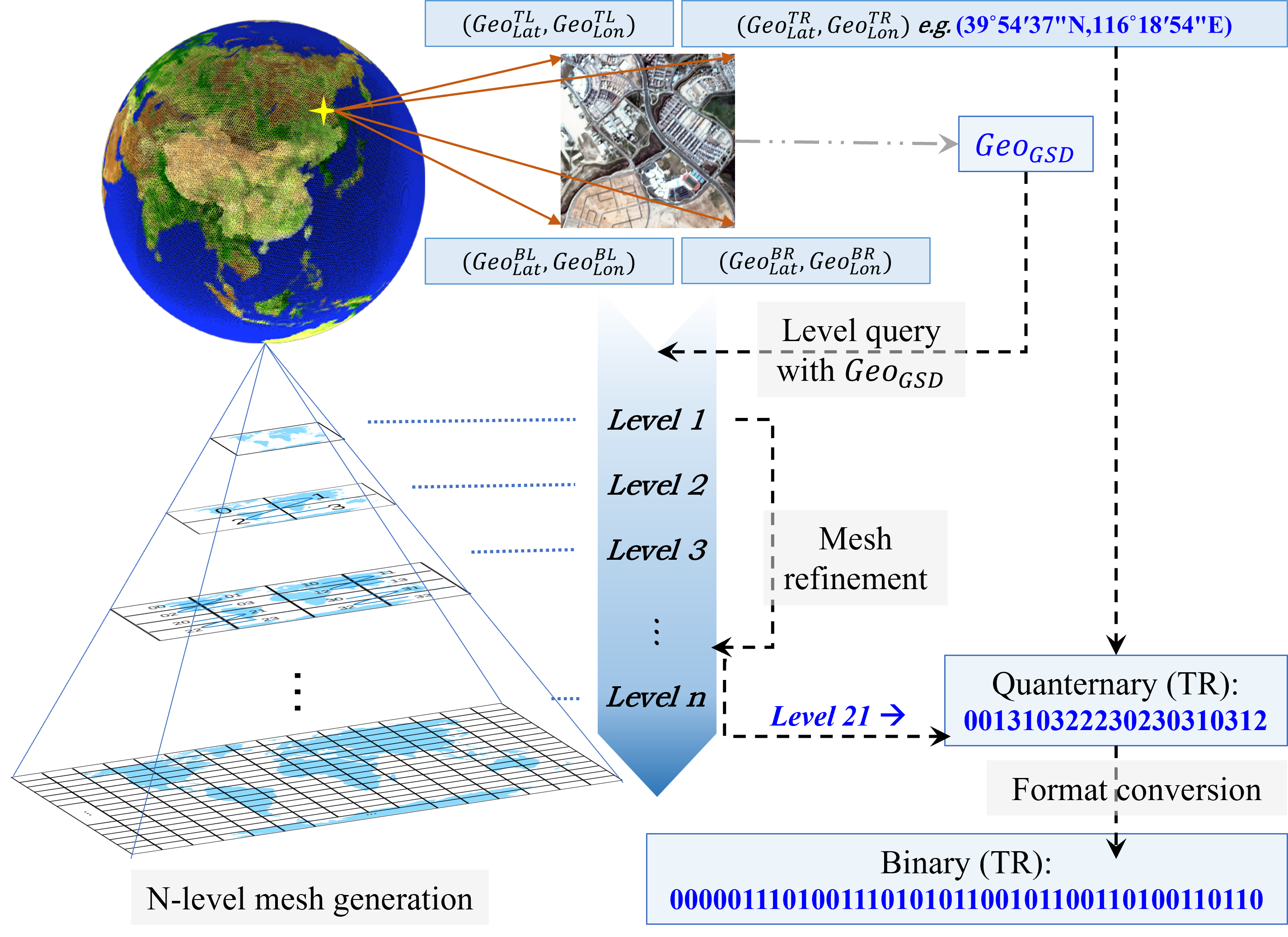}
    \caption{The proposed geographic encoding module in A$^{2}$-MAE. By efficiently encoding the geographic metadata (\ie, a $Geo_{GSD}$ and four sets of $(Geo_{Lat}^{c}, Geo_{Lon}^{c})$ in RS images, the geographic encoding module encourages A$^{2}$-MAE to be explicitly aware of this crucial geographic prior for varying geographic information in images.}
    \label{fig:geoscale}
    \vspace{-0.2cm}
\end{figure}
The metadata stored with the RS images contains geographic information, including the latitude, longitude, and GSD. The latitude and longitude information indicates the absolute location of the retrieved image on the Earth, which is of significance in leveraging the geographic pattern in the pre-training on worldwide RS datasets. The GSD indicates the ground scale of the RS image, which is critical to understanding the spatial ranges and frequency specificity of the image. For example, an image with a low GSD has more details in high frequency than a high GSD image does. Therefore, we propose the GEM to explicitly incorporate essential geographic priors into the MAE model, thereby enhancing its generalization capabilities for downstream applications.

As illustrated in Figure \ref{fig:geoscale}, given an RS image, the corresponding metadata records one GSD $Geo_{GSD}$ and four sets of latitude and longitude (\ie, the four corners $(Geo_{Lat}^{c}, Geo_{Lon}^{c}), c \in \{TL, TR, BL, BR\}$). Instead of directly utilizing the decimal geographic information, the GEM views the RS image as a group of squared grids and encodes it to achieve better representative geo-encoding features. Let $\mathbb {G}$ be the set of grids formed with latitudes and longitudes. The $\mathbb {G}$ contains several levels of mesh in a pyramid way, which are composed of equally subdivided grids with integer coding. Let \emph{Level 0} mesh $G_{0} = \{g_{0}\} \in \mathbb {G}$ be a $512^{\circ} \times 512^{\circ}$ grid which covers the whole global area. \emph{Level 1} mesh $G_{1} = \{g_{1}, ..., g_{n1}\} \in \mathbb {G}$ is defined as equally subdivided four grids, each of which has $256^{\circ} \times 256^{\circ}$ in the height and width. Following this logic, \emph{Level k} mesh $G_{k}$ is obtained by a quadtree division of \emph{Level k-1} mesh $G_{k-1}$. The higher level mesh represents finer resolution with low GSD. In all, given the metadata of an RS image, we first query the closest \emph{Level k} according to its $Geo_{GSD}$, then four sets of latitude and longitude are embedded as sequences of binary array. For example, The $Geo_{GSD}$ of Landsat-8 images is 30 m, which can be referred to grids in \emph{Level 21} (32 m of the size of each grid in the equator). Considering that the $Geo_{GSD}$ is utilized as an approximate version in this encoding strategy, we further encode the precise $Geo_{GSD}$ by replacing the positional embedded vector with the ground scaled positional encoding vector (inspired by \cite{reed2023scalemae}), which can be embedded as follows:

\begin{equation}
    v_{gsd,x}(pos, 2i) = sin \frac{\widehat{Geo_{GSD}}}{Geo_{GSD}} \frac{pos}{10000^{\frac{2i}{D}}},
\end{equation}

\begin{equation}
    v_{gsd,y}(pos, 2i+1) = cos \frac{\widehat{Geo_{GSD}}}{Geo_{GSD}} \frac{pos}{10000^{\frac{2i}{D}}},
\end{equation}

\noindent where $pos$ is the position of the embedded patch along the given axis, $i$ is the patch index, and $D$ is the number of embedded dimensions, exactly
 as introduced in \cite{vaswani2017attention}. $\widehat{Geo_{GSD}}$ is the reference GSD (nominally set to 1 m). 
 
 As a result, the proposed GEM efficiently embeds the geographic metadata, providing unique embedded features for images with specific locations and GSDs.

\section{Experiments}

\subsection{Implementation Details and Baselines}

We adopt ViT-Large architecture as the backbone for the proposed A$^{2}$-MAE, and pre-train the A$^{2}$-MAE using the constructed STSSD. We employ a progressive training strategy \cite{hong2024spectralgpt} by starting with S2-L8 data and then progressively transitioning to GF-S2 in STSSD. The patch size is fixed to $16\times16$ pixels. A$^{2}$-MAE is pre-trained for 130 epochs with a batch size of 1,024 on 8 NVIDIA A800 GPUs. AdamW optimizer \cite{he2016deep} is utilized with an initial learning rate of 0.0001, coupled with a half-cycle cosine decay schedule. According to existing works \cite{he2022masked,cong2022satmae}, we adopt a masking ratio of 75\%, balancing training efficiency and pretext task difficulty. Five self-supervised learning methods with the officially released pre-trained weights are selected as competing methods in this study, including 2 ResNet-50 based methods (SeCo \cite{manas2021seco}, CACo \cite{mall2023caco}) and 3 ViT-Large based methods (vanilla MAE pre-trained on ImageNet-1k~\cite{he2022masked}, SatMAE (the version for spectral data) \cite{cong2022satmae}, and ScaleMAE \cite{reed2023scalemae}). Full fine-tuning is employed in the downstream tasks for all methods, of which the first layer is modified to fit the data structure of specific datasets.

%The patch size is fixed at 16, independent of resolution, with each spectral band undergoing random masking. However, temporal encoding does not enhance performance on our dataset and is excluded from this study. MLP/UpperNet serve as decoders for segmentation/change detection tasks, and all methods undergo full fine-tuning for downstream tasks. Supplementary material provides details on epochs. For baselines, we use officially released weights for fine-tuning. In Strategy 2, we ensure spatial overlap among selected images by cropping $I_{i}^{t2, s2}$ within the geographical scope of $P_{i}^{t1,s1}$, which will be clarified in the final version.

\subsection{Comparison Results}\label{sec:com}

We conduct experiments on 7 datasets with diverse distributions in spatial, temporal, and spectral coverage, encompassing various data sources. This ensures a comprehensive assessment of the capabilities in efficiently utilizing spatial-, temporal-, and spectral-variant features, involving different downstream tasks, including classification, segmentation, and change detection. We employ the encoder of the competing pre-trained models for all downstream tasks, and details of the training setups for fine-tuning of each task are included in the supplementary materials. As shown in Table \ref{tab:compare}, A$^{2}$-MAE achieves comprehensive improvements across downstream tasks, indicating the effectiveness of A$^{2}$-MAE in exploiting the properties of RS images. 

\textbf{Land Cover Classification:} We perform the scene classification task on EuroSAT \cite{helber2019eurosat} and the multi-label classification task on BigEarthNet \cite{sumbul2019bigearthnet}. EuroSAT comprises 27K Sentinel-2 images with 13 bands collected from 34 European countries. BigEarthNet encompasses 590K Sentinel-2 images with 13 bands collected from 10 countries. As presented in Table \ref{tab:compare}, despite that several competing methods (\ie, SeCo, CACo, and SatMAE) are custom-tailored and pre-trained on Sentinel-2 image dataset, A$^{2}$-MAE still outperforms all competing methods in both EuroSAT and BigEarthNet datasets, highlighting the proposed A$^{2}$-MAE's effectiveness in leveraging diverse RS information within one unified model.

\begin{table}[]
\vspace{-0.1cm}
\caption{Comparison results (\%) of classification, segmentation, and change detection. The vanilla MAE model was pre-trained on ImageNet-1k.}
\scalebox{0.82}{
\begin{tabular}{c|cc|cc|ccc}
\midrule
\multirow{3}{*}{Methods} & \multicolumn{2}{c|}{Classification} & \multicolumn{2}{c|}{Segmentation}  & \multicolumn{3}{c}{Change detection} \\ \cline{2-8}
   & EuroSAT & BigEarthNet & Sen1Floods11 & CropSeg & LEVIR-CD & OSCD & DynamicEarthNet \\
                         & Accuracy        & mAP               & mIoU         & mIoU                & mIoU               & F1         & mIoU \\ \hline
SeCo\cite{manas2021seco} & 95.63           & 82.6              & 79.10       & 32.78                & 82.37              & 46.78     & 43.1    \\
CACo\cite{mall2023caco} & 95.90           & 82.1              & 84.62       & 32.83               & 82.90              & 51.74      & 41.2 \\ \hline
Vanilla MAE\cite{he2022masked} & 98.78      & 80.1              & 85.42        & 44.51               & 80.41              & 34.74    & 39.8  \\
SatMAE\cite{cong2022satmae}& 98.98           & 82.1              & 88.77        & 44.21               & 82.11              & 47.63     & 44.7 \\
ScaleMAE\cite{reed2023scalemae}& 98.31           & 82.2              & 85.05        & 44.04               & 83.07              & 48.70     & 44.3      \\
A2-MAE                   & \textbf{99.09}   & \textbf{83.0}    & \textbf{88.87}   & \textbf{44.81}   & \textbf{84.32}      & \textbf{53.97}  & \textbf{46.0}   \\ \midrule
\end{tabular}
}
\vspace{-0.1cm}
\label{tab:compare}
\end{table}

\textbf{Semantic Segmentation:}
We perform experiments on Sen1Floods11 \cite{bonafilia2020sen1floods11} and CropSeg\cite{Geospatial}. Sen1Floods11 is a surface water segmentation dataset including 4,831 Sentinel-2 imagery with 13 bands covering 120,406 km$^{2}$ and spans 14 biomes and 6 continents of the world across 11 flood events. CropSeg is a cropland segmentation dataset containing 3,854 Harmonized Landsat-Sentinel imagery with 7 bands at 30 m resolution across the Contiguous United States. Given that SeCo and CACo are pre-trained on datasets covering only urban regions, A$^{2}$-MAE, which is pre-trained on STSSD covering diverse land cover types, achieves significant improvements by 9.77\%/12.03\% against SeCo and 4.25\%/11.98\% against CACo, highlighting its superior generalization ability when pre-trained on STSSD with diverse coverage and geographical characteristics.

\textbf{Change Detection:}
We conduct experiments on the LEVIR-CD \cite{chen2020spatial}, OSCD \cite{daudt2018oscd}, and DynamicEarthNet datasets \cite{toker2022dynamicearthnet}. LEVIR-CD comprises 637 image pairs with a resolution of 0.5 m and a time span ranging from 5 to 14 years. The OSCD dataset comprises Sentinel-2 images with 13 bands collected from 24 urbanized regions worldwide. DynamicEarthNet provides daily images from Planet with 4 bands at 3 m resolution and monthly images from Sentinel-2 with 13 bands across approximately 75 areas of interest worldwide. Table \ref{tab:compare} presents the quantitative evaluation results of baselines and A$^{2}$-MAE. A$^{2}$-MAE outperforms competing methods across various backbones and self-supervised architectures by 1.25\% on mIoU / 2.23\% on F1 / 1.3\% on mIoU against the second-best results in LEVIR-CD, OSCD, and DynamicEarthNet, respectively. These observed improvements underscore the effectiveness of the proposed A$^{2}$-MAE in harnessing multi-temporal RS images within a single unified model.

\subsection{Ablation Study} \label{sec:ablation}
To efficiently and fairly investigate the key contributions of A$^{2}$-MAE, \ie, AAM and GEM, we conduct ablation experiments by pre-training and fine-tuning A$^{2}$-MAE on DynamicEarthNet, which is split into 55 locations for training and 10 locations for testing. The pre-training and the training phase of fine-tuning are conducted on the split training locations. A scratch version of SatMAE, which is pre-trained and further fine-tuned on DynamicEarthNet training sites, is viewed as the baseline. We also conduct comparisons in terms of masking strategy (\ie, random masking and tube masking strategies \cite{cong2022satmae}) and geographic encoding method (\ie, one-hot geographic encoding \cite{christie2018fmow} and scale encoding \cite{reed2023scalemae}). We further encode the geographic information using the proposed GEM during fine-tuning on DynamicEarthNet. This configuration (denoted as A$^{2}$-MAE$^{+}$) gives us a promising glance at the potential of fully utilizing the GEM when the geographic metadata of RS images is available in downstream RS tasks.

\begin{table}
\centering
\vspace{-0.1cm}
\caption{Performance comparison (\%) between different sub-components of A$^{2}$-MAE on DynamicEarthNet change detection. SatMAE* denotes a scratch version of SatMAE architecture pre-trained on DynamicEarthNet and further fine-tuned on DynamicEarthNet. A$^{2}$-MAE$^{+}$ denotes introducing the geographic encoding during fine-tuning.}
\scalebox{0.85}{
\begin{tabular}{lll}
\hline
Pre-training & Pix. Acc. & mIoU  \\ \hline
SatMAE* \cite{cong2022satmae}  & 72.2     & 45.9 \\ \hline
Van. MAE (random masking)    & 69.2     & 40.2 \\
+ tube masking \cite{cong2022satmae}    & 67.4 & 38.6 \\
+ AAM        & 72.7     & 46.8 \\ \hline
+ One-hot geo-encoding\cite{christie2018fmow}   & 71.9  &  47.1  \\ 
+ scale encoding \cite{reed2023scalemae}        & 72.8     & 47.3 \\
+ full GEM (A$^{2}$-MAE)  & 73.8     & 47.7 \\ \hline
A$^{2}$-MAE$^{+}$       & 76.4     & 56.1 \\ \hline
\end{tabular}
}
\label{tab:ablation}
\end{table}

As shown in Table \ref{tab:ablation}, A$^{2}$-MAE outperforms the scratched SatMAE by 1.6/0.8 of Pix. Acc./mIoU, decoupling and showcasing the contributions of the proposed pre-training method. Besides, both the AAM (+ 6.6\% on mIoU) and GEM (+ 0.9\% on mIoU) contribute to the significant performance improvements of A$^{2}$-MAE. Specifically, for masking strategies, the proposed AAM outperforms random masking and tube masking strategies by 6.6\% on mIoU, indicating the effectiveness of AAM. For geographic encoding methods, comparisons against one-hot geographic encoding \cite{christie2018fmow} show improvements by 1.9\%/0.6\% on Acc./mIoU. Further ablation studies reveal enhancements by 0.1\%/0.5\% on Acc./mIoU for GSD embedding and 1.0\%/0.4\% on Acc./mIoU for Lat/Lon embedding. 

Furthermore, by introducing the geographic information via the GEM in fine-tuning phase, A$^{2}$-MAE$^{+}$ achieves a notable improvement of 8.4\% on mIoU against A$^{2}$-MAE, indicating the promising potential of the GEM. It reveals that for downstream tasks that provide raw geographic metadata, introducing the GEM during fine-tuning can improve the results by a large margin.

%\section{Discussion}
\subsection{Discussion on Model Efficiency in Exploiting RS Images with Diverse Characteristics}

Given the vast quantity and varied characteristics of RS images, it is crucial to efficiently exploit the intrinsic relevance of images with diverse spatial, temporal, and spectral attributes. Previous studies have predominantly focused on addressing specific facets of this diversity, such as multi-spectral \cite{hong2024spectralgpt} and multi-resolution images \cite{reed2023scalemae}, or by achieving separate pre-training models \cite{cong2022satmae}. Consequently, achieving comprehensive and generalizable improvements across downstream tasks that span spatial, temporal, and spectral dimensions remains challenging. In response, a contemporaneous study, Skysense \cite{guo2023skysense}, designs separate backbones for three types of sources in a larger model with 2.06 billion parameters, which is trained on 80 A100 GPUs. However, this approach faces difficulties in scaling model parameters and computational overhead to accommodate expanding RS sources, which is evidently unsustainable. Different from this method, A$^{2}$-MAE explores the utilization of various multi-spectral sources in a unified backbone. Benefiting from the proposed anchor-aware masking strategy, A$^{2}$-MAE enables the efficient exploitation of the intrinsic complementarity information within RS images from different multi-spectral sources within one unified spatial-temporal-spectral model. Moreover, it requires 6$\times$ fewer model parameters than \cite{guo2023skysense}, thus facilitating efficient pre-training utilizing only 8 A800 GPUs, significantly saving computational costs while achieving comprehensive improvements across various downstream tasks.

%\subsection{Limitations and Future Work}

%Despite the global coverage provided by Sentinel and Landsat images, the 0.8 m high-resolution data in the STSSD dataset is currently limited to China due to constraints in data acquisition. In the future, expanding the volume of the STSSD is promising through the adoption of progressive training strategies \cite{hong2024spectralgpt} for efficient pre-training on large-scale datasets within the proposed unified model A$^{2}$-MAE. Additionally, future endeavors aim to integrate contrastive learning techniques \cite{he2020moco} and enhance the diversity of remote sensing (RS) data for pre-training by incorporating various modalities \cite{xiong2024all} such as Synthetic Aperture Radar and hyperspectral images.

\section{Conclusion}

In this study, we introduce STSSD, a spatial-temporal-spectral structured dataset comprising 510K sampling locations with 2.5 million structured images collected from multiple RS sources. To exploit different kinds of multi-spectral sources in one unified backbone, we propose an anchor-aware masking strategy to harness the intrinsic complementary information from different kinds of images, thus achieving more powerful feature representations. Furthermore, we propose the geographic encoding module to leverage geographic information, thereby improving the model generalization ability. Experiments verify the effectiveness and advantages of our method compared to existing RS pre-training models with the same parameter amount across image classification, semantic segmentation, and change detection tasks. In future work, we will expand the diversity of modalities such as Synthetic Aperture Radar and hyperspectral images in STSSD and A$^{2}$-MAE.
%Empirical evidence demonstrates that A$^{2}$-MAE, pre-trained on STSSD, achieves state-of-the-art performance on a broad range of downstream RS tasks, including image classification, semantic segmentation, and change detection tasks.

\clearpage  % TODO REVIEW/FINAL: This \clearpage needs to be removed from both review and camera-ready versions.

\bibliographystyle{splncs04}
\bibliography{main}

\begin{thebibliography}{10}
\providecommand{\url}[1]{\texttt{#1}}
\providecommand{\urlprefix}{URL }
\providecommand{\doi}[1]{https://doi.org/#1}

\bibitem{abu2016youtube}
Abu-El-Haija, S., Kothari, N., Lee, J., Natsev, P., Toderici, G., Varadarajan, B., Vijayanarasimhan, S.: Youtube-8m: A large-scale video classification benchmark. arXiv preprint arXiv:1609.08675  (2016)

\bibitem{ayush2021geography}
Ayush, K., Uzkent, B., Meng, C., Tanmay, K., Burke, M., Lobell, D., Ermon, S.: Geography-aware self-supervised learning. In: Proceedings of the IEEE/CVF International Conference on Computer Vision. pp. 10181--10190 (2021)

\bibitem{bachmann2022multimae}
Bachmann, R., Mizrahi, D., Atanov, A., Zamir, A.: Multimae: Multi-modal multi-task masked autoencoders. In: European Conference on Computer Vision. pp. 348--367. Springer (2022)

\bibitem{bai2023geographic}
Bai, L., Huang, W., Zhang, X., Du, S., Cong, G., Wang, H., Liu, B.: Geographic mapping with unsupervised multi-modal representation learning from vhr images and pois. ISPRS Journal of Photogrammetry and Remote Sensing  \textbf{201},  193--208 (2023)

\bibitem{bastani2023satlaspretrain}
Bastani, F., Wolters, P., Gupta, R., Ferdinando, J., Kembhavi, A.: Satlaspretrain: A large-scale dataset for remote sensing image understanding. In: Proceedings of the IEEE/CVF International Conference on Computer Vision. pp. 16772--16782 (2023)

\bibitem{bonafilia2020sen1floods11}
Bonafilia, D., Tellman, B., Anderson, T., Issenberg, E.: Sen1floods11: A georeferenced dataset to train and test deep learning flood algorithms for sentinel-1. In: Proceedings of the IEEE/CVF Conference on Computer Vision and Pattern Recognition Workshops. pp. 210--211 (2020)

\bibitem{caron2021emerging}
Caron, M., Touvron, H., Misra, I., J{\'e}gou, H., Mairal, J., Bojanowski, P., Joulin, A.: Emerging properties in self-supervised vision transformers. In: Proceedings of the IEEE/CVF international conference on computer vision. pp. 9650--9660 (2021)

\bibitem{chen2020spatial}
Chen, H., Shi, Z.: A spatial-temporal attention-based method and a new dataset for remote sensing image change detection. Remote Sensing  \textbf{12}(10), ~1662 (2020)

\bibitem{chen2020simple}
Chen, T., Kornblith, S., Norouzi, M., Hinton, G.: A simple framework for contrastive learning of visual representations. In: International conference on machine learning. pp. 1597--1607. PMLR (2020)

\bibitem{chen2021exploring}
Chen, X., He, K.: Exploring simple siamese representation learning. In: Proceedings of the IEEE/CVF conference on computer vision and pattern recognition. pp. 15750--15758 (2021)

\bibitem{christie2018fmow}
Christie, G., Fendley, N., Wilson, J., Mukherjee, R.: Functional map of the world. In: Proceedings of the IEEE Conference on Computer Vision and Pattern Recognition. pp. 6172--6180 (2018)

\bibitem{geoiccv}
Chu, G., Potetz, B., Wang, W., Howard, A., Song, Y., Brucher, F., Leung, T., Adam, H.: Geo-aware networks for fine-grained recognition. In: 2019 IEEE/CVF International Conference on Computer Vision Workshop (ICCVW). pp. 247--254 (2019)

\bibitem{cong2022satmae}
Cong, Y., Khanna, S., Meng, C., Liu, P., Rozi, E., He, Y., Burke, M., Lobell, D., Ermon, S.: Satmae: Pre-training transformers for temporal and multi-spectral satellite imagery. Advances in Neural Information Processing Systems  \textbf{35},  197--211 (2022)

\bibitem{daudt2018oscd}
Daudt, R.C., Le~Saux, B., Boulch, A., Gousseau, Y.: Urban change detection for multispectral earth observation using convolutional neural networks. In: IGARSS 2018-2018 IEEE International Geoscience and Remote Sensing Symposium. pp. 2115--2118. Ieee (2018)

\bibitem{deng2009imagenet}
Deng, J., Dong, W., Socher, R., Li, L.J., Li, K., Fei-Fei, L.: Imagenet: A large-scale hierarchical image database. In: 2009 IEEE conference on computer vision and pattern recognition. pp. 248--255. Ieee (2009)

\bibitem{everingham2015pascal}
Everingham, M., Eslami, S.A., Van~Gool, L., Williams, C.K., Winn, J., Zisserman, A.: The pascal visual object classes challenge: A retrospective. International journal of computer vision  \textbf{111},  98--136 (2015)

\bibitem{Geospatial}
Geospatial, I.N.: Ibm-nasa-geospatial/multi-temporal-crop-classification · datasets at hugging face, \url{https://huggingface.co/datasets/ibm-nasa-geospatial/multi-temporal-crop-classification}

\bibitem{grill2020bootstrap}
Grill, J.B., Strub, F., Altch{\'e}, F., Tallec, C., Richemond, P., Buchatskaya, E., Doersch, C., Avila~Pires, B., Guo, Z., Gheshlaghi~Azar, M., et~al.: Bootstrap your own latent-a new approach to self-supervised learning. Advances in neural information processing systems  \textbf{33},  21271--21284 (2020)

\bibitem{guo2023skysense}
Guo, X., Lao, J., Dang, B., Zhang, Y., Yu, L., Ru, L., Zhong, L., Huang, Z., Wu, K., Hu, D., He, H., Wang, J., Chen, J., Yang, M., Zhang, Y., Li, Y.: Skysense: A multi-modal remote sensing foundation model towards universal interpretation for earth observation imagery (2023)

\bibitem{he2022masked}
He, K., Chen, X., Xie, S., Li, Y., Doll{\'a}r, P., Girshick, R.: Masked autoencoders are scalable vision learners. In: Proceedings of the IEEE/CVF Conference on Computer Vision and Pattern Recognition. pp. 16000--16009 (2022)

\bibitem{he2020moco}
He, K., Fan, H., Wu, Y., Xie, S., Girshick, R.: Momentum contrast for unsupervised visual representation learning. In: Proceedings of the IEEE/CVF conference on computer vision and pattern recognition. pp. 9729--9738 (2020)

\bibitem{he2016deep}
He, K., Zhang, X., Ren, S., Sun, J.: Deep residual learning for image recognition. In: Proceedings of the IEEE Conference on Computer Vision and Pattern Recognition. pp. 770--778 (2016)

\bibitem{helber2019eurosat}
Helber, P., Bischke, B., Dengel, A., Borth, D.: Eurosat: A novel dataset and deep learning benchmark for land use and land cover classification. IEEE Journal of Selected Topics in Applied Earth Observations and Remote Sensing  \textbf{12}(7),  2217--2226 (2019)

\bibitem{hong2024spectralgpt}
Hong, D., Zhang, B., Li, X., Li, Y., Li, C., Yao, J., Yokoya, N., Li, H., Ghamisi, P., Jia, X., Plaza, A., Gamba, P., Benediktsson, J.A., Chanussot, J.: Spectralgpt: Spectral remote sensing foundation model (2024)

\bibitem{RN76}
Huang, X., Schneider, A., Friedl, M.A.: Mapping sub-pixel urban expansion in china using modis and dmsp/ols nighttime lights. Remote Sensing of Environment  \textbf{175},  92--108 (2016). \doi{10.1016/j.rse.2015.12.042}

\bibitem{RN75}
Imhoff, M.L., Lawrence, W.T., Elvidge, C.D., Paul, T., Levine, E., Privalsky, M.V., Brown, V.: Using nighttime dmsp/ols images of city lights to estimate the impact of urban land use on soil resources in the united states. Remote Sensing of Environment  \textbf{59}(1),  105--117 (1997). \doi{10.1016/S0034-4257(96)00110-1}

\bibitem{satclip}
Klemmer, K., Rolf, E., Robinson, C., Mackey, L., Rußwurm, M.: Satclip: Global, general-purpose location embeddings with satellite imagery (2023)

\bibitem{li2021geographical}
Li, W., Chen, K., Chen, H., Shi, Z.: Geographical knowledge-driven representation learning for remote sensing images. IEEE Transactions on Geoscience and Remote Sensing  \textbf{60},  1--16 (2021)

\bibitem{lin2014microsoft}
Lin, T.Y., Maire, M., Belongie, S., Hays, J., Perona, P., Ramanan, D., Doll{\'a}r, P., Zitnick, C.L.: Microsoft coco: Common objects in context. In: Computer Vision--ECCV 2014: 13th European Conference, Zurich, Switzerland, September 6-12, 2014, Proceedings, Part V 13. pp. 740--755. Springer (2014)

\bibitem{Long2021DiRS}
Long, Y., Xia, G.S., Li, S., Yang, W., Yang, M.Y., Zhu, X.X., Zhang, L., Li, D.: On creating benchmark dataset for aerial image interpretation: Reviews, guidances and million-aid. IEEE Journal of Selected Topics in Applied Earth Observations and Remote Sensing  \textbf{14},  4205--4230 (2021)

\bibitem{Long2022ASP}
Long, Y., Xia, G.S., Zhang, L., Cheng, G., Li, D.: Aerial scene parsing: From tile-level scene classification to pixel-wise semantic labeling (2022)

\bibitem{mall2023caco}
Mall, U., Hariharan, B., Bala, K.: Change-aware sampling and contrastive learning for satellite images. In: Proceedings of the IEEE/CVF Conference on Computer Vision and Pattern Recognition. pp. 5261--5270 (2023)

\bibitem{manas2021seco}
Manas, O., Lacoste, A., Gir{\'o}-i Nieto, X., Vazquez, D., Rodriguez, P.: Seasonal contrast: Unsupervised pre-training from uncurated remote sensing data. In: Proceedings of the IEEE/CVF International Conference on Computer Vision. pp. 9414--9423 (2021)

\bibitem{reed2023scalemae}
Reed, C.J., Gupta, R., Li, S., Brockman, S., Funk, C., Clipp, B., Keutzer, K., Candido, S., Uyttendaele, M., Darrell, T.: Scale-mae: A scale-aware masked autoencoder for multiscale geospatial representation learning. In: Proceedings of the IEEE/CVF International Conference on Computer Vision. pp. 4088--4099 (2023)

\bibitem{2023geographic}
Rußwurm, M., Klemmer, K., Rolf, E., Zbinden, R., Tuia, D.: Geographic location encoding with spherical harmonics and sinusoidal representation networks (2023)

\bibitem{grocvprw}
Scheibenreif, L., Hanna, J., Mommert, M., Borth, D.: Self-supervised vision transformers for land-cover segmentation and classification. In: 2022 IEEE/CVF Conference on Computer Vision and Pattern Recognition Workshops (CVPRW). pp. 1421--1430 (2022)

\bibitem{schmitt2019sen12ms}
Schmitt, M., Hughes, L.H., Qiu, C., Zhu, X.X.: Sen12ms--a curated dataset of georeferenced multi-spectral sentinel-1/2 imagery for deep learning and data fusion. arXiv preprint arXiv:1906.07789  (2019)

\bibitem{RN19}
Small, C., Pozzi, F., Elvidge, C.D.: Spatial analysis of global urban extent from dmsp-ols night lights. Remote Sensing of Environment  \textbf{96}(3-4),  277--291 (2005). \doi{10.1016/j.rse.2005.02.002}

\bibitem{sumbul2019bigearthnet}
Sumbul, G., Charfuelan, M., Demir, B., Markl, V.: Bigearthnet: A large-scale benchmark archive for remote sensing image understanding. In: IGARSS 2019-2019 IEEE International Geoscience and Remote Sensing Symposium. pp. 5901--5904. IEEE (2019)

\bibitem{sumbul2021bigearthnet}
Sumbul, G., De~Wall, A., Kreuziger, T., Marcelino, F., Costa, H., Benevides, P., Caetano, M., Demir, B., Markl, V.: Bigearthnet-mm: A large-scale, multimodal, multilabel benchmark archive for remote sensing image classification and retrieval [software and data sets]. IEEE Geoscience and Remote Sensing Magazine  \textbf{9}(3),  174--180 (2021)

\bibitem{toker2022dynamicearthnet}
Toker, A., Kondmann, L., Weber, M., Eisenberger, M., Camero, A., Hu, J., Hoderlein, A.P., {\c{S}}enaras, {\c{C}}., Davis, T., Cremers, D., et~al.: Dynamicearthnet: Daily multi-spectral satellite dataset for semantic change segmentation. In: Proceedings of the IEEE/CVF Conference on Computer Vision and Pattern Recognition. pp. 21158--21167 (2022)

\bibitem{tong2022videomae}
Tong, Z., Song, Y., Wang, J., Wang, L.: Videomae: Masked autoencoders are data-efficient learners for self-supervised video pre-training. Advances in neural information processing systems  \textbf{35},  10078--10093 (2022)

\bibitem{turner2014sensing}
Turner, W.: Sensing biodiversity. Science  \textbf{346}(6207),  301--302 (2014)

\bibitem{vaswani2017attention}
Vaswani, A., Shazeer, N., Parmar, N., Uszkoreit, J., Jones, L., Gomez, A.N., Kaiser, {\L}., Polosukhin, I.: Attention is all you need. Advances in neural information processing systems  \textbf{30} (2017)

\bibitem{wang2019remote}
Wang, R., Gamon, J.A.: Remote sensing of terrestrial plant biodiversity. Remote Sensing of Environment  \textbf{231},  111218 (2019)

\bibitem{wang2022self}
Wang, Y., Albrecht, C.M., Braham, N.A.A., Mou, L., Zhu, X.X.: Self-supervised learning in remote sensing: A review. arXiv preprint arXiv:2206.13188  (2022)

\bibitem{wang2022ssl4eo}
Wang, Y., Braham, N.A.A., Xiong, Z., Liu, C., Albrecht, C.M., Zhu, X.X.: Ssl4eo-s12: A large-scale multi-modal, multi-temporal dataset for self-supervised learning in earth observation. arXiv preprint arXiv:2211.07044  (2022)

\bibitem{woo2023convnext}
Woo, S., Debnath, S., Hu, R., Chen, X., Liu, Z., Kweon, I.S., Xie, S.: Convnext v2: Co-designing and scaling convnets with masked autoencoders. In: Proceedings of the IEEE/CVF Conference on Computer Vision and Pattern Recognition. pp. 16133--16142 (2023)

\bibitem{WU202187}
Wu, X., Shi, Z., Zou, Z.: A geographic information-driven method and a new large scale dataset for remote sensing cloud/snow detection. ISPRS Journal of Photogrammetry and Remote Sensing  \textbf{174},  87--104 (2021). \doi{https://doi.org/10.1016/j.isprsjprs.2021.01.023}, \url{https://www.sciencedirect.com/science/article/pii/S0924271621000290}

\bibitem{yao2023ringmo}
Yao, F., Lu, W., Yang, H., Xu, L., Liu, C., Hu, L., Yu, H., Liu, N., Deng, C., Tang, D., et~al.: Ringmo-sense: Remote sensing foundation model for spatiotemporal prediction via spatiotemporal evolution disentangling. IEEE Transactions on Geoscience and Remote Sensing  (2023)

\bibitem{zhang2022graph}
Zhang, S., Chen, H., Yang, H., Sun, X., Yu, P.S., Xu, G.: Graph masked autoencoders with transformers. arXiv preprint arXiv:2202.08391  (2022)

\bibitem{zhao2023mine}
Zhao, D., Wang, Q., Zhang, J., Bai, C.: Mine diversified contents of multi-spectral cloud images along with geographical information for multi-label classification. IEEE Transactions on Geoscience and Remote Sensing  (2023)

\bibitem{RN16}
Zhou, Y., Smith, S.J., Zhao, K., Imhoff, M., Thomson, A., Bond-Lamberty, B., Asrar, G.R., Zhang, X., He, C., Elvidge, C.D.: A global map of urban extent from nightlights. Environmental Research Letters  \textbf{10}(5),  054011 (2015). \doi{10.1088/1748-9326/10/5/054011}

\end{thebibliography}


\begin{thebibliography}{10}
\providecommand{\url}[1]{\texttt{#1}}
\providecommand{\urlprefix}{URL }
\providecommand{\doi}[1]{https://doi.org/#1}

\bibitem{bandara2022transformer}
Bandara, W.G.C., Patel, V.M.: A transformer-based siamese network for change detection. In: IGARSS 2022-2022 IEEE International Geoscience and Remote Sensing Symposium. pp. 207--210. IEEE (2022)

\bibitem{bastani2023satlaspretrain}
Bastani, F., Wolters, P., Gupta, R., Ferdinando, J., Kembhavi, A.: Satlaspretrain: A large-scale dataset for remote sensing image understanding. In: Proceedings of the IEEE/CVF International Conference on Computer Vision. pp. 16772--16782 (2023)

\bibitem{bingham2019sixty}
Bingham, H.C., Juffe~Bignoli, D., Lewis, E., MacSharry, B., Burgess, N.D., Visconti, P., Deguignet, M., Misrachi, M., Walpole, M., Stewart, J.L., et~al.: Sixty years of tracking conservation progress using the world database on protected areas. Nature ecology \& evolution  \textbf{3}(5),  737--743 (2019)

\bibitem{christie2018fmow}
Christie, G., Fendley, N., Wilson, J., Mukherjee, R.: Functional map of the world. In: Proceedings of the IEEE Conference on Computer Vision and Pattern Recognition. pp. 6172--6180 (2018)

\bibitem{cong2022satmae}
Cong, Y., Khanna, S., Meng, C., Liu, P., Rozi, E., He, Y., Burke, M., Lobell, D., Ermon, S.: Satmae: Pre-training transformers for temporal and multi-spectral satellite imagery. Advances in Neural Information Processing Systems  \textbf{35},  197--211 (2022)

\bibitem{elsen2018reply}
Elsen, P.R., Monahan, W.B., Merenlender, A.M.: Reply to you et al.: The world database on protected areas is an invaluable resource for global conservation assessments and planning. Proceedings of the National Academy of Sciences  \textbf{115}(39),  E9029--E9030 (2018)

\bibitem{he2022masked}
He, K., Chen, X., Xie, S., Li, Y., Doll{\'a}r, P., Girshick, R.: Masked autoencoders are scalable vision learners. In: Proceedings of the IEEE/CVF Conference on Computer Vision and Pattern Recognition. pp. 16000--16009 (2022)

\bibitem{mall2023caco}
Mall, U., Hariharan, B., Bala, K.: Change-aware sampling and contrastive learning for satellite images. In: Proceedings of the IEEE/CVF Conference on Computer Vision and Pattern Recognition. pp. 5261--5270 (2023)

\bibitem{manas2021seco}
Manas, O., Lacoste, A., Gir{\'o}-i Nieto, X., Vazquez, D., Rodriguez, P.: Seasonal contrast: Unsupervised pre-training from uncurated remote sensing data. In: Proceedings of the IEEE/CVF International Conference on Computer Vision. pp. 9414--9423 (2021)

\bibitem{oquab2024dinov2}
Oquab, M., Darcet, T., Moutakanni, T., Vo, H., Szafraniec, M., Khalidov, V., Fernandez, P., Haziza, D., Massa, F., El-Nouby, A., Assran, M., Ballas, N., Galuba, W., Howes, R., Huang, P.Y., Li, S.W., Misra, I., Rabbat, M., Sharma, V., Synnaeve, G., Xu, H., Jegou, H., Mairal, J., Labatut, P., Joulin, A., Bojanowski, P.: Dinov2: Learning robust visual features without supervision (2024)

\bibitem{reed2023scalemae}
Reed, C.J., Gupta, R., Li, S., Brockman, S., Funk, C., Clipp, B., Keutzer, K., Candido, S., Uyttendaele, M., Darrell, T.: Scale-mae: A scale-aware masked autoencoder for multiscale geospatial representation learning. In: Proceedings of the IEEE/CVF International Conference on Computer Vision. pp. 4088--4099 (2023)

\bibitem{sorscher2022beyond}
Sorscher, B., Geirhos, R., Shekhar, S., Ganguli, S., Morcos, A.: Beyond neural scaling laws: beating power law scaling via data pruning. Advances in Neural Information Processing Systems  \textbf{35},  19523--19536 (2022)

\bibitem{sumbul2019bigearthnet}
Sumbul, G., Charfuelan, M., Demir, B., Markl, V.: Bigearthnet: A large-scale benchmark archive for remote sensing image understanding. In: IGARSS 2019-2019 IEEE International Geoscience and Remote Sensing Symposium. pp. 5901--5904. IEEE (2019)

\bibitem{sumbul2021bigearthnet}
Sumbul, G., De~Wall, A., Kreuziger, T., Marcelino, F., Costa, H., Benevides, P., Caetano, M., Demir, B., Markl, V.: Bigearthnet-mm: A large-scale, multimodal, multilabel benchmark archive for remote sensing image classification and retrieval [software and data sets]. IEEE Geoscience and Remote Sensing Magazine  \textbf{9}(3),  174--180 (2021)

\bibitem{toker2022dynamicearthnet}
Toker, A., Kondmann, L., Weber, M., Eisenberger, M., Camero, A., Hu, J., Hoderlein, A.P., {\c{S}}enaras, {\c{C}}., Davis, T., Cremers, D., et~al.: Dynamicearthnet: Daily multi-spectral satellite dataset for semantic change segmentation. In: Proceedings of the IEEE/CVF Conference on Computer Vision and Pattern Recognition. pp. 21158--21167 (2022)

\bibitem{wang2022ssl4eo}
Wang, Y., Braham, N.A.A., Xiong, Z., Liu, C., Albrecht, C.M., Zhu, X.X.: Ssl4eo-s12: A large-scale multi-modal, multi-temporal dataset for self-supervised learning in earth observation. arXiv preprint arXiv:2211.07044  (2022)

\end{thebibliography}
\end{document}

\title{Supplementary material for A$^{2}$-MAE: A spatial-temporal-spectral unified remote sensing pre-training method based on anchor-aware masked autoencoder} 

\titlerunning{Abbreviated paper title}

\author{Lixian Zhang\inst{1,2,\thanks{Those authors contribute equally to this work.}} \and Yi Zhao\inst{2,\footnotemark[1]} \and Runmin Dong\inst{2, \footnotemark[1], \thanks{Corresponding authors.}} \and Jinxiao Zhang\inst{2} \and Shuai Yuan\inst{3} \and Shilei Cao\inst{4} \and Mengxuan Chen\inst{2} \and Juepeng Zheng\inst{4,\footnotemark[2]} \and Weijia Li\inst{4} \and Wayne Zhang\inst{5} \and Wei Liu\inst{5} \and Litong Feng\inst{5} \and Haohuan Fu\inst{2,\footnotemark[2]}}

\authorrunning{Zhang, L. et al.}

\institute{$^{1}$National Supercomputing Center in Shenzhen \quad
$^{2}$Tsinghua University \quad \\
$^{3}$The University of Hong Kong \quad
$^{4}$Sun Yat-Sen University \quad
$^{5}$SenseTime Group
{\tt\small zhanglx18@tsinghua.org.cn}~
{\tt\small \{drm,haohuan\}@mail.tsinghua.edu.cn}
}

\maketitle

\section{Overview}

The supplementary material elaborates on the details of our dataset, offers more details provides more details on the implementation of downstream tasks, and presents additional experimental results. The structure is outlined as follows:
%In this supplementary material, we offer additional content to enhance the comprehension of our paper, providing a nuanced understanding of the subject matter beyond the scope of the main text.

\begin{itemize}[label=\textbullet, leftmargin=2em]

\item We augment the understanding of the constructed dataset STSSD in Section \ref{sec:STSSD}. Specifically, in Section \ref{sec:datacomp}, we summarize the comparison between the public large-scale RS dataset and STSSD. In Section \ref{sec:databand}, detailed information regarding the spectral bands of the RS sources utilized in STSSD (Sentinel-2, Landsat-8, Gaofen-1, and Gaofen-2) is provided. In Section \ref{sec:dataid}, the identification of nature reserves and cities in STSSD construction is detailed. In Section \ref{sec:dataprun}, the data pruning strategy used in STSSD construction is presented, offering insights into STSSD's heterogeneity. Section \ref{sec:datacon} outlines the implementation details and hyperparameters used in STSSD construction.

\item Further implementation details regarding the configuration of fine-tuning stages and additional evaluation metrics for downstream tasks are provided in Section \ref{sec:train}. Specifically, in Section \ref{sec:traincomp}, a comparison of all utilized downstream datasets and additional implementation details for fine-tuning stages setup are presented. Section \ref{sec:comdino} compares our model with a randomly initialized counterpart and the representative Vision Foundation Model, DINOv2. Comprehensive evaluation metrics for utilized downstream tasks are provided in Section \ref{sec:trainmetrics}. Section \ref{sec:vis} offers visualization comparisons in downstream tasks.

\item More experiments are provided to supplement the analysis of the proposed A$^{2}$-MAE, improving understanding of its effectiveness in Section \ref{sec:ablation}. Specifically, in Section \ref{sec:ablmask}, the results of hyperparameter tuning of the masking ratio in A$^{2}$-MAE are presented. Section \ref{sec:ablfro} evaluates model performance on a downstream task with frozen backbone features.

\end{itemize}

\section{Addition information regarding STSSD}\label{sec:STSSD}

\subsection{Comparison with public RS dataset}\label{sec:datacomp}
Since researchers have introduced several large-scale RS datasets \cite{christie2018fmow,sumbul2019bigearthnet,sumbul2021bigearthnet,toker2022dynamicearthnet,wang2022ssl4eo,bastani2023satlaspretrain}, we provide a comprehensive comparison of STSSD against existing public datasets in remote sensing in Table \ref{dataset}. 

Our STSSD dataset offers several notable contributions and strengths compared to existing public datasets in remote sensing domain. Firstly, STSSD boasts a comprehensive spatial coverage, spanning globally, and a relatively recent temporal range from 2020 to 2023. This temporal range surpasses many other datasets, including fMoW\cite{christie2018fmow}, BigEarthNet\cite{sumbul2019bigearthnet}, and DynamicEarthNet\cite{toker2022dynamicearthnet}. Moreover, STSSD provides a diverse range of ground sampling distances (GSD) ranging from 0.8 to 30 meters, offering finer spatial resolutions than most datasets, such as BigEarthNet and SSL4EO-S12\cite{wang2022ssl4eo}. Additionally, STSSD incorporates data from four distinct sources, surpassing the number of sources in other datasets like BigEarthNet, BigEarthNet-MM, and DynamicEarthNet. Consequently, STSSD stands out for its extensive coverage, recent temporal span, multiple spatial resolutions, and diverse data sources, collectively contributing to its significance as a valuable resource for remote sensing applications.

\begin{table*}
\centering
\caption{Comparison of STSSD against existing public datasets in remote sensing (K=thousands).}
\begin{tabular}{lccccc}
\hline
DataSet         & Year & Spatial cover & Temporal spanning & GSD (m)                       & \#sources   \\ \hline
fMoW \cite{christie2018fmow}           & 2018 & global        &  2002-2017                 & 0.31-1.60            & 1                                         \\
BigEarthNet \cite{sumbul2019bigearthnet}    & 2019 & Europe        & 2017-2018         & 10                            & 1                                         \\
BigEarthNet-MM \cite{sumbul2021bigearthnet}  & 2021 & Europe        & 2017-2018         & 10                        & 2                                        \\
DynamicEarthNet \cite{toker2022dynamicearthnet} & 2022 & global        & 2018-2019         & 3.0                           & 2                                            \\
SSL4EO-S12 \cite{wang2022ssl4eo}     & 2023 & global        & 2021              & 10                          & 2                                       \\
SatlasPretrain \cite{bastani2023satlaspretrain} & 2023 & global        & 2011-2022         & 1-10                 & 3                                       \\
STSSD            & 2023 & global        & 2020-2023         & 0.8-30 & 4     \\ \hline
\end{tabular}
\label{dataset}
\end{table*}

\subsection{Spectral bands in STSSD}\label{sec:databand}

\begin{table*}[htbp]
\centering
\caption{The statistics and spectral information of the used Sentinel-2 in the STSSD. Columns 5 and 6 (\ie, Mean and Standard deviation) represent the mean and standard deviation of pixel values for each channel across the retrieved Sentinel-2 dataset, respectively. We resize the bands of B1, B5-7, and B8a-12 to 10 m, achieving harmonious images with all 12-band of 10 m.}
\begin{tabular}{clcccc}
\hline
Satellite                    & Channel       & Resolution & Central wavelength & Mean              & Standard deviation \\ \hline
\multirow{12}{*}{Sentinel-2} & B1: Aerosols   & 60 m        & 443 nm             & 1161.52          & 523.98           \\
                             & B2: Blue       & 10 m        & 490 nm             & 1399.38          & 536.77           \\
                             & B3: Green      & 10 m        & 560 nm             & 1455.72          & 625.56           \\
                             & B4: Red        & 10 m        & 665 nm             & 2761.06          & 771.05           \\
                             & B5: Red Edge1   & 20 m        & 705 nm             & 1815.21          & 616.86          \\
                             & B6: Red Edge2   & 20 m        & 740 nm             & 2465.56          & 671.86          \\
                             & B7: Red Edge3   & 20 m        & 783 nm             & 2722.34          & 727.05 \\
                             & B8: NIR        & 10 m        & 842 nm             & 2867.82          & 756.91          \\
                             & B8a: Red Edge4  & 20 m        & 865 nm             & 2336.82          & 683.80           \\
                             & B9: Water Vapor & 60 m        & 940 nm             & 1742.15           & 629.27           \\
                             & B11: SWIR1     & 20 m        & 1610 nm            & 1069.34          & 465.96          \\
                             & B12: SWIR2     & 20 m        & 2190 nm            & 3128.77          & 848.05          \\ \hline
\end{tabular}
\label{tab:sen2}
\end{table*}

\begin{table*}[htbp]
\centering
\caption{The statistics and spectral information of used Landsat-8 in the STSSD. Columns 5 and 6 (\ie, Mean and Standard deviation) represent the mean and standard deviation of pixel values for each channel across the retrieved Landsat-8 dataset, respectively.}
\begin{tabular}{clcccc}
\hline
Satellite                  & Channel    & Resolution & Central wavelength & Mean     & Standard deviation \\ \hline
\multirow{7}{*}{Landsat-8} & B1: Coastal & 30 m         & 440 nm              & 229.17 & 262.79           \\
                           & B2: Blue       & 30 m         & 480 nm              & 277.28 & 266.60           \\
                           & B3: Green      & 30 m         & 560 nm              & 488.15 & 277.24           \\
                           & B4: Red        & 30 m         & 655 nm              & 411.19 & 305.21           \\
                           & B5: NIR        & 30 m         & 865 nm              & 2104.24 & 612.09          \\
                           & B6: SWIR 1     & 30 m         & 1610 nm             & 1275.84 & 452.19          \\
                           & B7: SWIR 2     & 30 m         & 2200 nm             & 701.26 & 325.94          \\ \hline
\end{tabular}
\label{tab:lan8}
\end{table*}

\begin{table*}[htbp]
\centering
\caption{The raw resolution, statistic, and spectral information of used Gaofen-1 and Gaofen-2 in the STSSD. Columns 5 and 6 (\ie, Mean and Standard deviation) represent the mean and standard deviation of pixel values for each channel across the retrieved Gaofen-1 and Gaofen-2 datasets, respectively. We perform the pan-sharpening technique to generate images with 2 m of all 4 bands of Gaofen-1 and 0.8 m of all 4 bands of Gaofen-2 imagery.}
\begin{tabular}{clcccc}
\hline
Satellite             & Channel & Resolution & Central wavelength & Mean     & Standard deviation \\ \hline
\multirow{4}{*}{Gaofen-1} & B1: Blue    & 8 m          & 485 nm              & 96.11 & 37.65           \\
                      & B2: Green   & 8 m          & 555 nm              & 98.68 & 36.20           \\
                      & B3: Red     & 8 m          & 660 nm              & 99.79 & 35.67           \\
                      & B4: NIR     & 8 m          & 830 nm              & 99.67 & 38.72           \\ \hline
\multirow{4}{*}{Gaofen-2} & B1: Blue    & 2 m        & 485 nm              & 78.67 & 33.51          \\
                      & B2: Green   & 2 m        & 555 nm              & 81.17 &  33.19     \\
                      & B3: Red     & 2 m        & 660 nm              & 85.88 & 33.76       \\
                      & B4: NIR     & 2 m        & 830 nm              & 86.31 & 36.01       \\ \hline
\end{tabular}
\label{tab:gf}
\end{table*}

Tables \ref{tab:sen2},  \ref{tab:lan8}, and  \ref{tab:gf} encapsulate a comprehensive summary of the spectral bands, along with their pixel-wise mean and standard deviation values across each spectral band for the four RS data sources utilized in the STSSD.

\subsection{The identification of nature reserves and cities in STSSD}\label{sec:dataid}

\textbf{City}: Initially, we employ a sampling strategy similar to that used for city sampling \cite{mall2023caco, manas2021seco}, with larger scales and finer filters. Cities are ranked based on the population size. The top 12,000 cities are selected for sampling. Subsequently, a Gaussian sampling approach is employed within a region centered at the geographical coordinates of each selected city, with a radius of 50 km. Upon selecting a candidate point (\ie, longitude, latitude) within this region, additional checks are conducted to ensure the spatial diversity (no marine areas, no overlapping areas), cloud cover ($\leq$ 10\% covering), etc. 

\textbf{Nature reserves}: For nature reserves, a similar approach are adopted, focusing on regions designated as protected natural areas. Geographic datasets containing information about nature reserves \cite{bingham2019sixty, elsen2018reply} are used to identify suitable sampling locations. The same sampling methodology described for cities is applied within the boundaries of these nature reserves, ensuring spatial coverage across diverse ecological habitats. Similarly, checks for cloud cover and proximity to previously sampled locations are conducted to ensure the integrity and representativeness of the sampled data.

\subsection{The implementation details regarding data pruning in STSSD}\label{sec:dataprun}

Due to the extensive scale of our dataset, challenging examples contribute significantly to its diversity and complexity, thus enhancing our model's feature representation capability. We hypothesize that images collected from urban areas are more complex and heterogeneous, whereas those from nature reserves tend to be redundant and homogeneous. Hence, we exclusively apply the data pruning strategy to the observation data from nature reserves. Initially, we partition the collected RS dataset from nature reserves into 341 non-overlapping sub-regions based on geo-locations, each delineated by a $1^{\circ}\times1^{\circ}$ grid. As geographical representative features within each sub-region vary significantly, we employ individual data pruning models based on k-means clustering to effectively preserve the geographical diversity within nature reserves.

\begin{figure*}
    \centering
    \vspace{-0.1cm}
    \includegraphics[width=\linewidth]{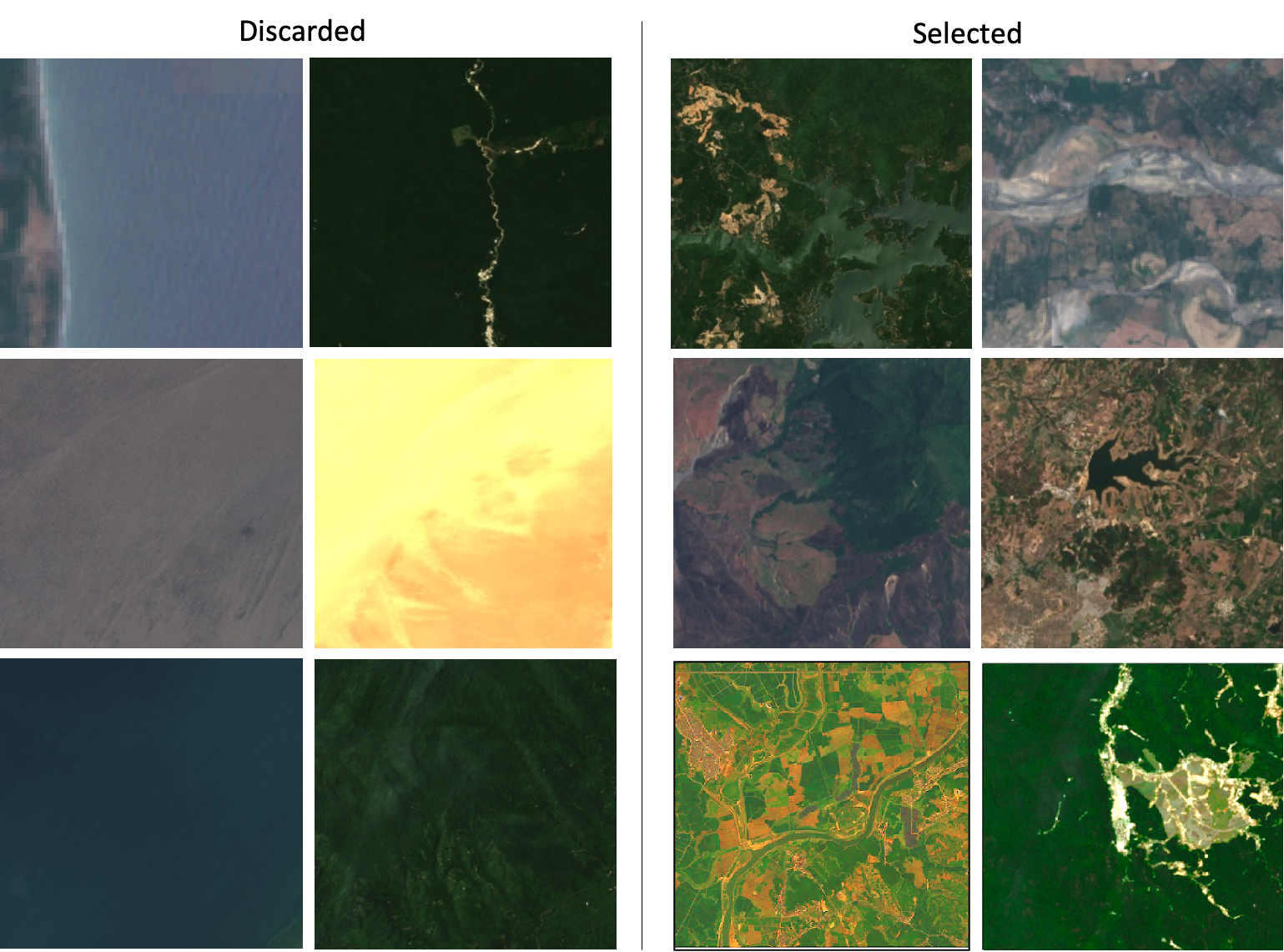}
    \caption{The visualization of discarded images and selected images in the data pruning strategy.}
    \label{fig:pruning}
    \vspace{-0.5cm}
\end{figure*}

In the data pruning approach, we gauge the difficulty of each data point by measuring its Euclidean distance to the nearest cluster centroid \cite{sorscher2022beyond}. Simple examples are those most prototypical, while hard examples deviate furthest from the norm. The simplest examples provide limited information, while the hard examples provide diverse information about earth observation. Specifically, adopting the common land cover classification system (IGBP), we set k, the number of clusters, to 20. After the k-means clustering, we remain the most difficult 10\% images of nature reserves in the final STSSD. To provide deeper insights into example difficulty across various metrics, we visually represent selected images and discarded images based on our self-supervised prototype metric (Fig. \ref{fig:pruning}). Qualitatively, simple examples typically exhibit high similarity and redundancy, while hard examples tend to high heterogeneous and complexity.

\subsection{The construction details of STSSD}\label{sec:datacon}

To enhance the quality of the constructed STSSD, targeted pre-processing steps are undertaken for each data source. Initially, all data undergoes processing for atmosphere and radiation correction. Subsequently, the Gaofen series images are pan-sharpened to achieve higher resolution.

Afterwards, the STSSD is organized based on geographical locations, with each location containing 3 to 6 pairs of images, contingent upon the data source availability in the respective geographical area. To ensure alignment across locations and address variations in image resolutions within each pair, the images are cropped to sizes ranging from 256 $\times$ 256 to 3,200 $\times$ 3,200. Specifically, for nature reserves, Landsat-8 images are resized to 256 $\times$ 256, while Sentinel-2 images are resized to 768 $\times$ 768. In the case of the main cities, Landsat-8 images are adjusted to 256 $\times$ 256, and Sentinel-2 images to 480 $\times$ 480. Moreover, for the geographic region of China, Sentinel-2 images are standardized to 256 $\times$ 256, Gaofen-1 images to 1,280 $\times$ 1,280, and Gaofen-2 images to 3,200 $\times$ 3,200.

\begin{table}
\caption{Details of the datasets utilized in downstream tasks. }
\resizebox{\linewidth}{!}{
\begin{tabular}{llllll}
\hline
Dataset      & Spatial coverage                     & Temporal coverage & Spectral coverage                      & Resolution \\ \hline
EuroSAT      & 34 European countries                &  single year    & 13 bands of Sentinel-2 & 10 m              \\ \hline
BigEarthNet  & 10 countries                         &  2 years      & 13 bands of Sentinel-2 & 10 m              \\ \hline
Sen1Floods11 & 14 biomes &  single year      & 13 bands of Sentinel-2 & 10 m              \\ \hline
CropSeg      & Contiguous United States  &   single year         &  7 bands of HLS                         & 30 m              \\ \hline
LEVIR-CD     & 20 districts      & 5-14 years        & RGB                                    & 0.5 m             \\ \hline
OSCD         & 24 urbanized regions       &   3 years     & 13 bands of Sentinel-2  & 10 m              \\ \hline
\end{tabular}
}
\label{tab:coverage}
\end{table}

\section{Additional details regarding experiments on downstream tasks} \label{sec:train}

In this section, we elaborate on the training configurations utilized for fine-tuning models in downstream tasks in the main paper.

\subsection{Implementation details in downstream tasks}\label{sec:traincomp}

Full fine-tuning is employed in the downstream tasks for all methods, of which the first layer is modified to fit the data structure of specific downstream datasets. All competing methods adopt the same head for each downstream dataset (specified in Table \ref{tab:details}).

As presented in Table \ref{tab:coverage}, this study utilizes a range of downstream datasets and implements various tasks to evaluate the effectiveness of the proposed approach. To ensure a fair comparison, the competing methods adopt the officially released pre-trained weights for each downstream task. The utilized datasets include EuroSAT covering 34 European countries for a single year, BigEarthNet spanning 10 countries over two years, Sen1Floods11 representing 14 biomes in a single year, CropSeg focusing on the contiguous United States for a single year, LEVIR-CD covering 20 districts over 5-14 years, and OSCD encompassing 24 urbanized regions over three years. All the utilized downstream tasks contribute to a comprehensive evaluation of the generalization capability of the competing methods in addressing the diversities in spatial, temporal, and spectral coverage. All the competing methods are trained and evaluated on the same partition of training and test materials for each downstream task.

For the implementation details, as summarized in Table \ref{tab:details}, each dataset is associated with specific implementation details, including the input size, input channel, optimizer (AdamW), learning rate, learning rate schedule (multi-step), weight decay, batch size, maximum iterations/epoch, warmup strategy (linear), and task-specific head and loss function. 
%The experiments are designed to evaluate the performance of the proposed method across diverse geographical and temporal contexts, providing insights into its generalizability and effectiveness across different remote sensing tasks.

\begin{table}
\vspace{-0.3cm}
\caption{Implementation details in the downstream task fine-tuning. LR. indicates the learning rate. Sch. indicates the scheduler. Iter. indicates the iteration.}
\resizebox{\linewidth}{!}{
\begin{tabular}{l|llllll}
\midrule
Dataset         & EuroSAT           & BigEarthNet             & Sen1Floods11 & CropSeg    & LEVIR-CD                                                    & OSCD       \\ \midrule
Optimizer       & AdamW    & AdamW & AdamW & AdamW   & AdamW & AdamW            \\ \hline
Input size      & 96×96             & 128×128                 & 512×512      &   512×512         & 256×256                                                     & 96×96      \\ \hline
Input channel   & 13 bands     & 13 bands           & 6 bands          & 6 bands    & RGB                                                         & RGB        \\ \hline
Base LR.        & 1e-3             & 1e-3                   & 1e-4      &    1e-4        & 3e-4     &   3e-4    \\ \hline
LR. sch.        & multi-step & multi-step & multi-step & multi-step & multi-step & multi-step\\ \hline
Weight decay    & 0.01 & 0.01 & 0.01 & 0.01 & 0.01 & 0.01  \\ \hline
Batch size      &    128               & 256                     & 32           &    16        & 96  &   96 \\ \hline
Max epoch & 100 epoch         & 100 epoch               & 100 epoch  &   100 epoch         & 200 epoch                                                   & 100 epoch  \\ \hline
Warmup          & linear & linear & linear & linear & linear & linear     \\ \hline
Head            & Linear cls. & Linear cls.       & UperNet      & UperNet    & ChangeFormer \cite{bandara2022transformer} & U-Net      \\ \hline
Loss function   & CrossEntropy      & Soft-margin & BCE          & BCE        & BCE                                                         & BCE        \\ \midrule
\end{tabular}
}
\vspace{-0.3cm}
\label{tab:details}
\end{table}

\begin{figure*}[]
    \centering
    \includegraphics[width=10cm]{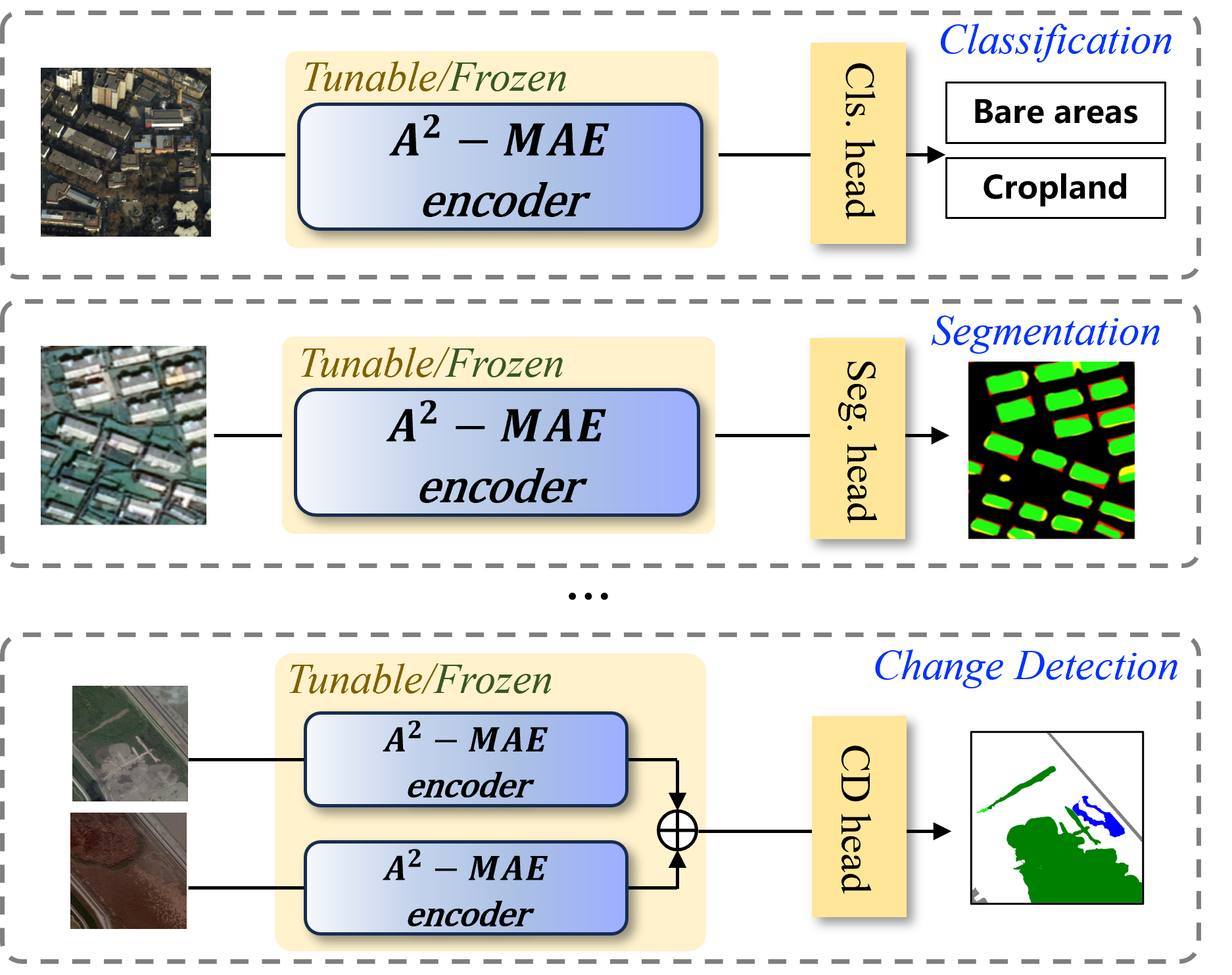}
    \caption{Illustration of the proposed A$^{2}$-MAE to accommodate different downstream tasks. A$^{2}$-MAE is flexible in addressing downstream tasks with different input combinations.}
    \label{fig:down}
    \vspace{-0.3cm}
\end{figure*}

The proposed A$^{2}$-MAE approach exhibits versatility in accommodating various downstream tasks, as depicted in Fig. \ref{fig:down}. By leveraging its flexibility, A$^{2}$-MAE can effectively address tasks such as scene classification, semantic segmentation, and change detection, each with distinct input combinations in terms of spectral, temporal, and spatial resolutions. This flexibility allows A$^{2}$-MAE to adapt to the specific requirements of each task, thereby enhancing its applicability across a wide range of remote sensing applications. Whether the task demands spectral information, temporal sequences, or high-resolution spatial details, A$^{2}$-MAE can seamlessly integrate these input modalities to deliver accurate and robust results. Consequently, the proposed approach presents a versatile solution for remote sensing practitioners, enabling them to tackle diverse challenges with a unified framework.

\subsection{Comparison with vision foundation model DINOv2 on downstream tasks} \label{sec:comdino}

The proposed A$^{2}$-MAE approach demonstrates superior performance compared to competing vision foundation models across various remote sensing tasks, as illustrated in Table \ref{tab:dino}. In classification tasks on EuroSAT and BigEarthNet datasets, A$^{2}$-MAE achieves the highest accuracy, outperforming the Vanilla MAE model pre-trained on ImageNet-1k. Similarly, in segmentation tasks on Sen1Floods11 and CropSeg datasets, A$^{2}$-MAE achieves the highest mIoU scores, indicating superior segmentation quality compared to both Vanilla MAE \cite{he2022masked} and DINOv2 \cite{oquab2024dinov2}. Moreover, in change detection tasks on LEVIR-CD and OSCD datasets, A$^{2}$-MAE consistently outperforms competing models in terms of mIoU and F1 scores, demonstrating its effectiveness in detecting changes accurately and comprehensively. These comprehensive evaluation results underscore the superior performance of the proposed A$^{2}$-MAE approach across multiple remote sensing tasks, highlighting its potential as a robust solution for various remote sensing applications.

\begin{table}
\centering
\vspace{-0.3cm}
\caption{Comparison results (\%) in classification, segmentation, and change detection tasks. The vanilla MAE model is pre-trained on ImageNet-1k.}
\scalebox{0.82}{
\begin{tabular}{c|cc|cc|cc}
\midrule
\multirow{3}{*}{Methods} & \multicolumn{2}{c|}{Classification} & \multicolumn{2}{c|}{Segmentation}  & \multicolumn{2}{c}{Change detection} \\ \cline{2-7}
   & EuroSAT & BigEarthNet & Sen1Floods11 & CropSeg & LEVIR-CD & OSCD \\
                         & Accuracy        & mAP               & mIoU         & mIoU                & mIoU               & F1 \\ \hline
Vanilla MAE \cite{he2022masked} & 98.78      & 80.1              & 85.42        & 44.51               & 80.41              & 34.74    \\
DINOv2 \cite{oquab2024dinov2}&  98.67          & 81.2              & 88.02         & 42.07               & 82.11              &  53.61 \\
A$^{2}$-MAE (ours)                   & \textbf{99.09}   & \textbf{83.0}    & \textbf{88.87}   & \textbf{44.81}   & \textbf{84.32}      & \textbf{53.97}  \\ \midrule
\end{tabular}
}
\vspace{-0.3cm}
\label{tab:dino}
\end{table}

\subsection{Comprehensive evaluation for the downstream tasks} \label{sec:trainmetrics}

As presented in Table \ref{tab:change} and Table \ref{tab:oscd}, the comprehensive evaluation of downstream tasks in this study, particularly change detection on LEVIR-CD and OSCD datasets, demonstrates the superiority of the proposed A$^{2}$-MAE method across various evaluation metrics compared to competing methods. On LEVIR-CD, A$^{2}$-MAE consistently outperforms other methods across different backbone architectures and self-supervised approaches, achieving the highest scores in F1, IoU, Recall, and Precision metrics. Notably, compared to ScaleMAE, the closest competitor, A$^{2}$-MAE exhibits a substantial improvement in IoU and Recall, indicating its effectiveness in detecting changes accurately. Similarly, on the OSCD dataset, A$^{2}$-MAE surpasses competing methods across all evaluation metrics, including F1, Precision, and Recall. Particularly noteworthy is the significant margin by which A$^{2}$-MAE outperforms other methods in terms of Recall, indicating its ability to capture a larger proportion of true change instances. 
%These results underscore the robustness and efficacy of the proposed A$^{2}$-MAE method for change detection tasks, highlighting its potential for enhancing remote sensing applications.

\begin{table}
\centering
\caption{Change detection results (\%) on LEVIR-CD. Sup.(ImageNet) indicates a supervised model pre-trained on ImageNet-1k.}
\begin{tabular}{llllll}
\hline
Method         & Backbone & F1             & IoU            & Recall         & Prec.          \\ \hline
SeCo\cite{manas2021seco}           & ResNet-50 & 90.33          & 82.37          & 88.91          & 91.82          \\
CACo\cite{mall2023caco}           & ResNet-50 & 90.56          & 82.90          & 89.31          & 92.04          \\ \hline
Sup.(ImageNet) & ViT-Large    & 90.14          & 80.41          & 88.17          & 91.72          \\
SatMAE\cite{cong2022satmae}         & ViT-Large    & 90.60          & 82.11          & 88.12          & 92.34          \\
ScaleMAE\cite{reed2023scalemae}       & ViT-Large    & 90.75          & 83.07          & 89.18          & \textbf{92.38} \\
A$^{2}$-MAE (ours)          & ViT-Large    & \textbf{91.09} & \textbf{84.32} & \textbf{90.39} & 92.34          \\ \hline
\end{tabular}
\label{tab:change}
\end{table}

\begin{table}
\centering
\caption{Change detection results (\%) on OSCD. Sup. (Scratch) indicates a supervised model trained from scratch (a randomly initialized network).}
\begin{tabular}{lllll}
\hline
Method   & Backbone & F1             & Precision      & Recall         \\ \hline
Sup. (Scratch) & ResNet-50 & 35.13          & 69.04          & 23.44          \\
SeCo~\cite{manas2021seco}     & ResNet-50 & 46.78          & 64.57          & 36.60          \\
CACo~\cite{mall2023caco}     & ResNet-50 & 51.74          & 62.38          & 44.78          \\
Sup.(Scratch) & ViT-Large    & 34.74          & 67.83          & 23.26          \\
ScaleMAE~\cite{reed2023scalemae} & ViT-Large    & 47.63          & 46.28          & 49.06          \\
SatMAE~\cite{cong2022satmae}   & ViT-Large    & 48.70          & 46.95          & 50.58          \\
A$^{2}$-MAE (ours)      & ViT-Large    & \textbf{53.97} & \textbf{50.93} & \textbf{57.40} \\ \hline
\end{tabular}
\label{tab:oscd}
\end{table}

\subsection{Visualization comparisons in the downstream task} \label{sec:vis}

\begin{figure*}[]
    \centering
    \includegraphics[width=\linewidth]{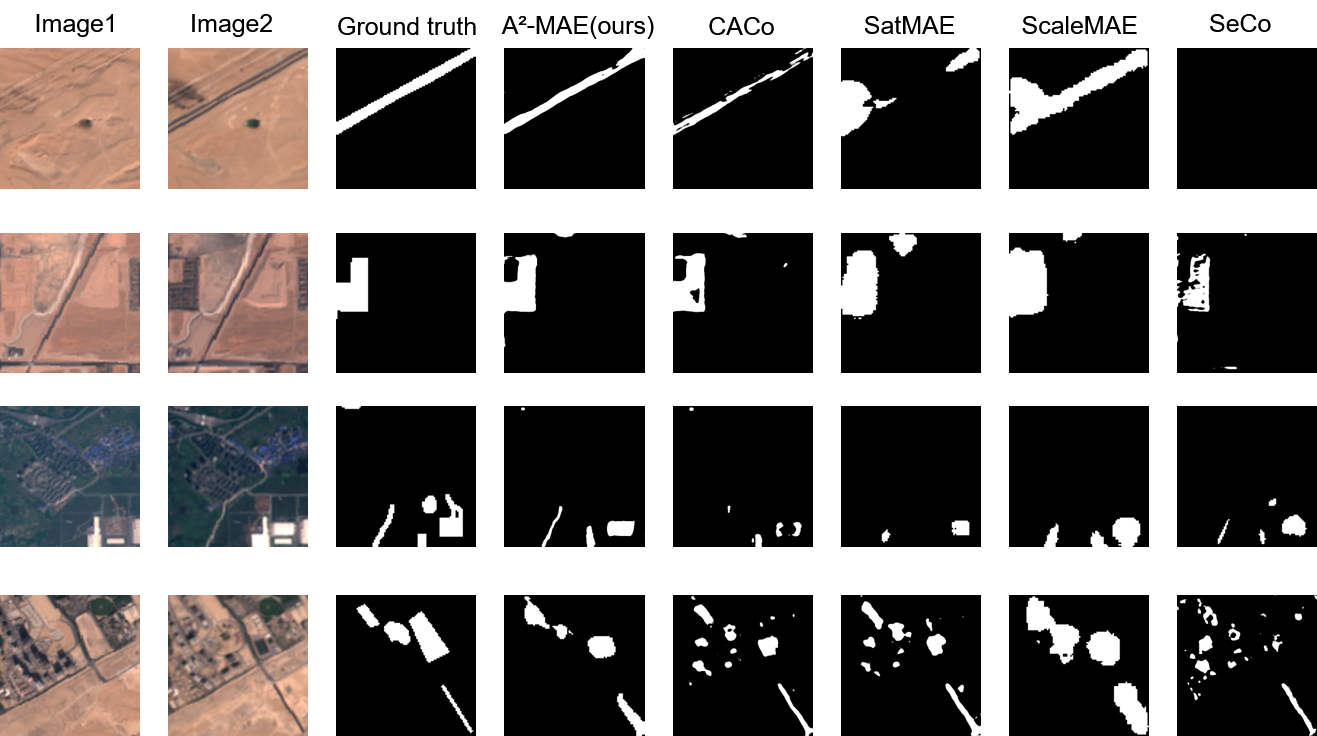}
    \caption{Visualization of change detection results on OSCD change detection. The columns from left to right are input images (Image 1 and Image 2), ground truth labels, and results of A$^{2}$-MAE (ours), CACo~\cite{mall2023caco}, SatMAE~\cite{cong2022satmae}, ScaleMAE~\cite{reed2023scalemae}, and SeCo~\cite{manas2021seco}, respectively.}
    \label{fig:change}
\vspace{-0.3cm}
\end{figure*}

In this section, we present cases of visualization comparisons in the downstream task, \ie, OSCD change detection. The visual comparisons of downstream task in Fig. \ref{fig:change} clearly demonstrate the superior performance of the proposed A$^{2}$-MAE method compared to competing methods. A$^{2}$-MAE exhibits superior performance in accurately detecting changes between two images, as indicated by its results being more closely aligned with the ground truth labels compared to CACo, SatMAE, ScaleMAE, and SeCo. These visual comparisons underscore the effectiveness of A$^{2}$-MAE in producing high-quality outputs, highlighting its potential for advancing the remote sensing applications.

\section{Additional experiments}\label{sec:ablation}

\subsection{Hyperparameter tuning of masking ratio}\label{sec:ablmask}

In this section, we conduct experiments on masking ratio tuning and outline the experimental results in Table \ref{tab:mask}. By varying the masking ratios to 0.6 and 0.9 from the default of 0.75, the impact on the mIoU is observed. A lower masking ratio of 0.6 leads to a slight decrease in mIoU by 0.2\%, indicating that reconstruction becomes somewhat easier due to the increased visibility of patches. Besides, a higher masking ratio of 0.9 results in a more significant decrease in mIoU by 0.5\%, suggesting a more challenging pretext task due to fewer visible patches. Consequently, a masking ratio of 0.75 is considered approximately optimal as it strikes a balance between ease of reconstruction and the difficulty of the pretext task, resulting in improved performance compared to the other two masking ratios.

\begin{table}
\centering
\vspace{-0.3cm}
\caption{The results (\%) of masking ratio tuning on SpaceNet dataset.}
\begin{tabular}{lll}
\hline
Making ratio & Backbone & mIoU \\ \hline
0.6 & ViT-Large & 79.5 \\
0.75 & ViT-Large & 79.7 \\
0.9 & ViT-Large & 79.2 \\ \hline

\end{tabular}
\label{tab:mask}
\vspace{-0.3cm}
\end{table}

\subsection{Results of linear probing in downstream tasks}\label{sec:ablfro}

The effectiveness of the A$^{2}$-MAE approach in leveraging pre-trained features with frozen backbones for downstream tasks involving multi-spectral images is demonstrated through experiments conducted on the BigEarthNet dataset. Specifically, A$^{2}$-MAE, pre-trained on datasets encompassing diverse spectral compositions, exhibits its adaptability to both 3-band (RGB) and multiple-band inputs. To comprehensively evaluate A$^{2}$-MAE's representative capabilities on BigEarthNet, we provide two versions of linear probing results. 

\begin{figure*}
    \centering
    \vspace{-0.3cm}
    \includegraphics[width=\linewidth]{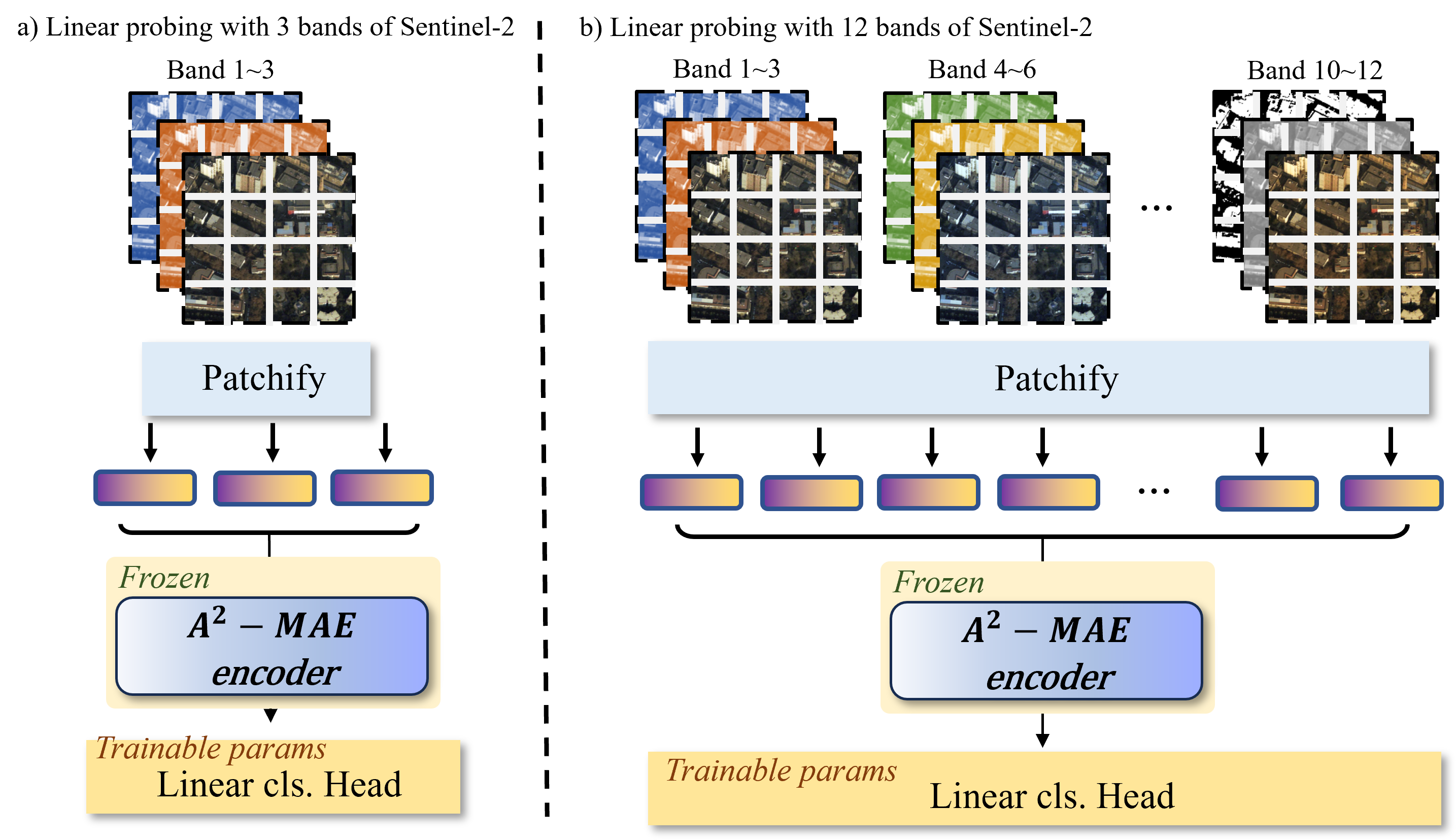}
    \caption{Illustration of the usage of A$^{2}$-MAE in a) 3 bands of Sentinel-2 and b) 12 bands of Sentinel-2 images with the frozen backbone.}
    \label{fig:freeze}
    \vspace{-0.3cm}
\end{figure*}

Firstly, we evaluate A$^{2}$-MAE utilizing only three bands (RGB) of Sentinel-2 images (Fig. \ref{fig:freeze} a), enabling a fair comparison against competing methods trained solely on 3-band imagery. Secondly, we explore A$^{2}$-MAE's potential by incorporating all 12 bands of Sentinel-2 (Fig. \ref{fig:freeze} b), showcasing its ability to fully unlock pre-trained representative features. In the latter case, we partition the input multi-spectral image into sub-images, each comprising three bands, and concatenate the tokens generated from these patches. This innovative process allows the frozen model not only to handle the standard 3-band Sentinel-2 images but also the full 12-band of spectral information.

The comparison results presented in Table \ref{tab:bigearthnet} underscore A$^{2}$-MAE's superior performance in multi-label classification tasks using images with 3-band of images on BigEarthNet, achieving a mean Average Precision (mAP) of 73.5\%. A$^{2}$-MAE outperforms competing models tailored for the standard 3-band images, including CACo \cite{mall2023caco}, SatMAE \cite{cong2022satmae}, and ScaleMAE \cite{reed2023scalemae}. Notably, A$^{2}$-MAE's utilization of frozen backbone features extends beyond the conventional RGB channels to encompass a broader spectrum (all 12-band of spectral information) with a mAP of 74.3\%, achieving 0.8\% higher than the 3-band version. The gained improvement is higher than that of SatMAE (+0.3\% of mAP when utilizing all 12-band of Sentinel-2 images). This capability enhances its adaptability to multi-spectral imagery, emphasizing the robustness and versatility of A$^{2}$-MAE in effectively harnessing images with diverse spectral combinations using the pre-trained features. 

\begin{table}
\centering
\vspace{-0.3cm}
\caption{Multi-label classification results (\%) on BigEarthNet. The mark of * indicates the version of the frozen model that adopts all 12-band of Sentinel-2 images.}
\begin{tabular}{llll}
\hline
Model & Input channels & Backbone & mAP \\ \hline
CACo \cite{mall2023caco} & B2,B3,B4 (RGB) & ResNet-50 & 72.6 \\ 
SatMAE \cite{cong2022satmae} & B2,B3,B4 (RGB) & ViT-Large & 66.2 \\
ScaleMAE \cite{reed2023scalemae} & B2,B3,B4 (RGB) & ViT-Large & 73.2 \\
A$^{2}$-MAE (ours) & B2,B3,B4 (RGB)  & ViT-Large & 73.5 \\ \hline
SatMAE* \cite{cong2022satmae} & B1-8,8A,9,11-12 & ViT-Large & 66.5 \\
A$^{2}$-MAE* (ours) & B1-8,8A,9,11-12 & ViT-Large & \textbf{74.3} \\ \hline
\end{tabular}
\vspace{-0.3cm}
\label{tab:bigearthnet}
\end{table}

\clearpage

\bibliographystyle{splncs04}
\bibliography{main}